\documentclass[runningheads]{llncs}
\usepackage{graphicx}

\usepackage{tikz}
\usepackage{comment}
\usepackage{amsmath,amssymb} 
\usepackage{color}

\usepackage{amsmath}
\usepackage{amssymb}
\usepackage{booktabs}
\usepackage{multirow}
\usepackage{graphicx}
\usepackage{caption}
\usepackage{subcaption}
\usepackage[ruled,linesnumbered]{algorithm2e}
\usepackage{rotating}
\usepackage{optidef}

\usepackage{array, makecell} %
\usepackage{bm}
\usepackage{booktabs, multicol, multirow}
\usepackage{hyperref}
\usepackage{diagbox}

\usepackage[accsupp]{axessibility}  

\begin{document}
\pagestyle{headings}
\mainmatter
\def\ECCVSubNumber{2598}  

\title{Depth Field Networks for Generalizable \\ Multi-view Scene Representation}

\newcommand{\MethodName}[0]{Depth Field Networks\xspace} 
\newcommand{\MethodAcronym}[0]{DeFiNe\xspace} 

\titlerunning{Depth Field Networks}
%
\author{Vitor Guizilini\inst{1}* \and
Igor Vasiljevic\inst{1}* \and
Jiading Fang\inst{2}* \and
Rares Ambrus\inst{1} \and 
Greg Shakhnarovich\inst{2} \and
Matthew R.\ Walter\inst{2} \and
Adrien Gaidon\inst{1}
}

\authorrunning{Guizilini et al.}
%
\institute{Toyota Research Institute, Los Altos, CA \and
Toyota Technological Institute at Chicago, Chicago, IL
}
\maketitle

\begin{figure}[h!]
\vspace{-8mm}
\centering
\subfloat{
\includegraphics[width=0.8\textwidth,height=3.23cm]{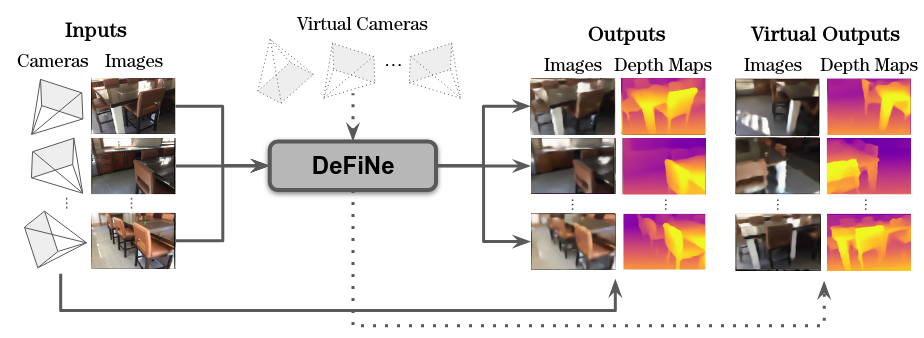}
}
\vspace{-3mm}
\caption{Our \textbf{\MethodName (\MethodAcronym)} achieve state of the art in multi-view depth estimation, while also enabling predictions from arbitrary viewpoints.%
}
\label{fig:teaser}
\vspace{-10mm}
\end{figure}

\begin{abstract}

Modern 3D computer vision leverages learning to boost geometric reasoning, mapping image data to classical structures such as cost volumes or epipolar constraints to improve matching. 
These architectures are specialized according to the particular problem, and thus require significant task-specific tuning, often leading to poor domain generalization performance.
Recently, generalist Transformer architectures have achieved impressive results in tasks such as optical flow and depth estimation by encoding geometric priors as inputs rather than as enforced constraints.
In this paper, we extend this idea and propose to learn an implicit, multi-view consistent scene representation, introducing a series of 3D data augmentation techniques as a geometric inductive prior to increase view diversity.  We also show that introducing view synthesis as an auxiliary task further improves depth estimation.
Our \MethodName (\MethodAcronym) achieve state-of-the-art results in stereo and video depth estimation without explicit geometric constraints, and improve on zero-shot domain generalization by a wide margin. Project page: \url{https://sites.google.com/view/tri-define}.
\let\thefootnote\relax\footnotetext{* Denotes equal contribution.}

\vspace{-2mm}

\end{abstract}

\section{Introduction}

Estimating 3D structure from a pair of images is a cornerstone problem of computer vision. Traditionally, this is treated as a correspondence problem, whereby one applies a homography to stereo-rectify the images and then matches pixels (or patches) along epipolar lines to obtain disparity estimates. 
Contemporary approaches to stereo are specialized variants of classical methods, relying on correspondences to compute cost volumes, epipolar losses, bundle adjustment objectives, or projective multi-view constraints, among others. These are either baked directly into the model architecture or enforced as part of the loss function.

Applying the principles of classical vision in this way has given rise to architectures that achieve state-of-the-art results on tasks such as stereo depth estimation~\cite{kendall2017end,lipson2021raft}, optical flow~\cite{raft}, and multi-view depth~\cite{deepv2d}. However, this success comes at a cost: each architecture is specialized and purpose-built for a single task. 
Great strides have been made to alleviate the dependence on strong geometric assumptions~\cite{gordon2019depth,vasiljevic2020neural}, and two recent trends allow us to \textit{decouple} the task from the architecture: (i) implicit representations and (ii) generalist networks.  Our work draws upon both of these directions. 

Implicit representations of geometry and coordinate-based networks have recently achieved incredible popularity in the vision community. This direction is pioneered by work on neural radiance fields (NeRF)~\cite{mildenhall2020nerf,xie2021neural}, where a point- and ray-based parameterization along with a volume rendering objective allow simple MLP-based networks to achieve state-of-the-art view synthesis results.  Follow-up works extend this coordinate-based representation to the pixel domain~\cite{pixelnerf}, allowing predicted views to be conditioned on image features. 
%
%
The second trend in computer vision has been the use of generalist architectures, pioneered by Vision Transformers~\cite{dosovitskiy2020image}. Emerging as an attention-based architecture for NLP, Transformers have been used for a diverse set of tasks, including depth estimation~\cite{li2021revisiting,ranftl2021vision}, optical flow~\cite{jaegle2021perceiverio}, and image generation~\cite{esser2021taming}. Transformers have also been applied to geometry-free view synthesis~\cite{rombach2021geometry}, demonstrating that attention can learn long-range correspondence between views for 2D-3D tasks. 
Representation Transformers (SRT)~\cite{sajjadi2021scene} use the Transformer encoder-decoder model to learn scene representations for view synthesis from sparse, high-baseline data with no geometric constraints.  However, owing to the quadratic scaling of the self-attention module, experiments are limited to low-resolution images and require very long training periods. 

To alleviate the quadratic complexity of self-attention, the Perceiver architecture~\cite{jaegle2021perceiver} disentangles the dimensionality of the latent representation from that of the inputs by fixing the size of the latent representation. Perceiver~IO~\cite{jaegle2021perceiverio} extends this architecture to allow for arbitrary outputs, with results on optical flow estimation that outperform traditional cost-volume based methods. Similarly, the recent Input-level Inductive Bias (IIB) architecture 
\cite{yifan2021input} uses image features and camera information as input to Perceiver~IO to directly regress stereo depth, outperforming baselines that use explicit geometric constraints. 
%
%
Building upon these works, we propose to learn a \textit{geometric scene representation} for depth synthesis from novel viewpoints, including estimation, interpolation, and extrapolation. We expand the IIB framework to the scene representation setting, taking sequences of images and predicting a consistent multi-view latent representation suitable for different downstream tasks.  
Taking advantage of the query-based nature of the Perceiver~IO architecture, we propose a series of 3D augmentations that increase viewpoint density and diversity during training, thus encouraging (rather than enforcing) multi-view consistency. Furthermore, we show that the introduction of view synthesis as an auxiliary task, decoded from the same latent representation, improves depth estimation without additional ground truth.

We test our model on the popular ScanNet benchmark~\cite{dai2017scannet}, achieving state-of-the-art real-time results for stereo depth estimation and competitive results for video depth estimation, without relying on memory- or compute-intensive operations such as cost volume aggregation and test-time optimization.  We show that our 3D augmentations lead to significant improvements over baselines that are limited to the viewpoint diversity of training data.
Furthermore, our zero-shot transfer results from ScanNet to 7-Scenes~\cite{shotton2013scene} improve the state-of-the-art by a large margin, demonstrating that our method generalizes better than specialized architectures, which suffer from poor performance on out-of-domain data.
Our contributions are summarized as follows:
\begin{itemize}
\item We use a generalist Transformer-based architecture to learn a depth estimator from an arbitrary number of posed images. In this setting, we (i) \textbf{propose a series of 3D augmentations} that improve the geometric consistency of our learned latent representation; and (ii) show that \textbf{jointly learning view synthesis as an auxiliary task improves depth estimation}.
\item Our \MethodName (\MethodAcronym) not only achieve \textbf{state-of-the-art stereo depth estimation results} on the widely used ScanNet dataset, but also exhibit superior generalization properties with \textbf{state-of-the-art results on zero-shot transfer to 7-Scenes.}
\item \MethodAcronym also \textbf{enables depth estimation from arbitrary viewpoints}.  We evaluate this novel generalization capability in the context of \emph{interpolation} (between timesteps), and \emph{extrapolation} (future timesteps).
\end{itemize}

\section{Related Work}
\subsubsection{Monocular Depth Estimation.}
Supervised depth estimation---the task of estimating per-pixel depth given an RGB image and a corresponding ground-truth depth map---dates back to the pioneering work of Saxena et al.~\cite{saxena2005learning}. 
Since then, deep learning-based architectures designed for supervised monocular depth estimation have become increasingly sophisticated~\cite{eigen2015predicting,eigen2014depth,fu2018deep,laina2016deeper,lee2019big}, generally offering improvements over the standard encoder-decoder convolutional architecture. 
Self-supervised methods provide an alternative to those that rely on ground-truth depth maps at training time, and are able to take advantage of the new availability of large-scale video datasets. Early self-supervised methods relied on stereo data~\cite{godard2017unsupervised}, and then progressed to fully monocular video sequences~\cite{zhou2017unsupervised}, with increasingly sophisticated losses~\cite{shu2020feature} and architectures~\cite{monodepth2,packnet,watson2021temporal}.


\vspace{-5mm}
\subsubsection{Multi-view Stereo.}

Traditional multi-view stereo approaches have dominated even in the deep learning era. COLMAP~\cite{schonberger2016structure} remains the standard framework for structure-from-motion, incorporating sophisticated bundle adjustment and keypoint refinement procedures, at the cost of speed. With the goal of producing closer to real-time estimates, multi-view stereo learning approaches adapt traditional cost volume-based approaches to stereo~\cite{kendall2017end,chang2018pyramid} and multi-view~\cite{yao2018mvsnet,im2019dpsnet} depth estimation, often relying on known extrinsics to warp views into the frame of the reference camera. 
Recently, iterative refinement approaches that employ recurrent neural networks have made impressive strides in optical flow estimation~\cite{raft}. Follow-on work applies this general recurrent correspondence architecture to stereo depth~\cite{lipson2021raft}, scene-flow~\cite{teed2021raft}, and even SLAM~\cite{teed2021droid}.  While their results are impressive, recurrent neural networks can be difficult to train, and test-time optimization increases inference time over a single forward pass.

Recently, Transformer-based architectures~\cite{attention_all} have replaced CNNs in many geometric estimation tasks.  The Stereo Transformer~\cite{li2021revisiting} architecture replaces cost volumes with a correspondence approach inspired by sequence-to-sequence modeling.
The Perceiver~IO~\cite{jaegle2021perceiverio} architecture constitutes a large departure from cost volumes and geometric losses. For the task of optical flow, Perceiver~IO feeds positionally encoded images through a Transformer~\cite{jaegle2021perceiver}, rather than using a cost volume for processing.  
IIB~\cite{yifan2021input} adapts the Perceiver~IO architecture to generalized stereo estimation, proposing a novel epipolar parameterization as an additional input-level inductive bias.  Building upon this baseline, 
we propose a series of geometry-preserving 3D data augmentation techniques designed to promote the learning of a \emph{geometrically-consistent latent scene representation}. We also introduce novel view synthesis as an auxiliary task to depth estimation, decoded from the same latent space. 
Our video-based representation (aided by our 3D augmentations) allows us to generalize to novel viewpoints, rather than be restricted to the stereo setting. 

\vspace{-5mm}
\subsubsection{Video Depth Estimation.}

Video and stereo depth estimation methods generally produce monocular depth estimates at test time. 
ManyDepth~\cite{watson2021temporal} combines a monocular depth framework with multi-view stereo, aggregating predictions in a cost volume and thus enabling multi-frame inference at test-time.
Recent methods accumulate predictions at train and test time, either with generalized stereo~\cite{ummenhofer2017demon} or with sequence data~\cite{zhou2018deeptam}.  DeepV2D~\cite{deepv2d} incorporates a cost-volume based multi-view stereo approach with an incremental pose estimator to iteratively improve depth and pose estimates at train and test time. 

Another line of work draws on the availability of monocular depth networks that perform accurate but \textit{multi-view inconsistent} estimates at test time~\cite{luo2020consistent}. In this setting, additional geometric constraints are enforced to finetune the network and improve multi-view consistency through epipolar constraints. 
Consistent Video Depth Estimation~\cite{luo2020consistent} refines COLMAP~\cite{schonberger2016structure} results with a monocular depth network constrained to be multi-view consistent. Subsequent work jointly optimizes depth and pose for added robustness to challenging scenarios with poor calibration~\cite{kopf2021robust}. A recent  framework incorporates many architectural elements of prior work into a Transformer-based architecture that takes video data as input for multi-view depth~\cite{long2021multi}. 
NeuralRecon~\cite{Sun_2021_CVPR} moves beyond depth-based architectures to learn Truncated Signed Distance Field (TSDF) volumes as a way to improve surface consistency. 

\vspace{-5mm}
\subsubsection{Novel View Synthesis.}
Since the emergence of neural radiance fields~\cite{mildenhall2020nerf}, implicit representations and volume rendering have emerged as the \textit{de facto} standard for view synthesis.  They parameterize viewing rays and points with positional encoding,
and need to be re-trained on a scene-by-scene basis.  Many recent improvements leverage depth supervision to improve view synthesis in a volume rendering framework~\cite{azinovic2021neural,deng2021depth,rematas2021urban,nerfingmvs,zhu2021nice}.
An alternative approach replaces volume rendering with a directly learned light field network~\cite{sitzmann2021light}, predicting color values directly from viewing rays. This is the approach we take when estimating the auxiliary view synthesis loss, due to its computational simplicity. 
Other works attempt to extend the NeRF approach beyond single scene models by incorporating learned~\textit{features}, enabling few-shot volume rendering~\cite{pixelnerf}.  Feature-based methods have also treated view synthesis as a sequence-learning task, such as the Scene Representation Transformer (SRT) architecture~\cite{sajjadi2021scene}.  

\section{The \MethodAcronym Architecture}




\begin{figure}[t!]
\centering
\subfloat[Architecture overview.]{
\label{fig:diagram}
\includegraphics[width=0.95\textwidth]{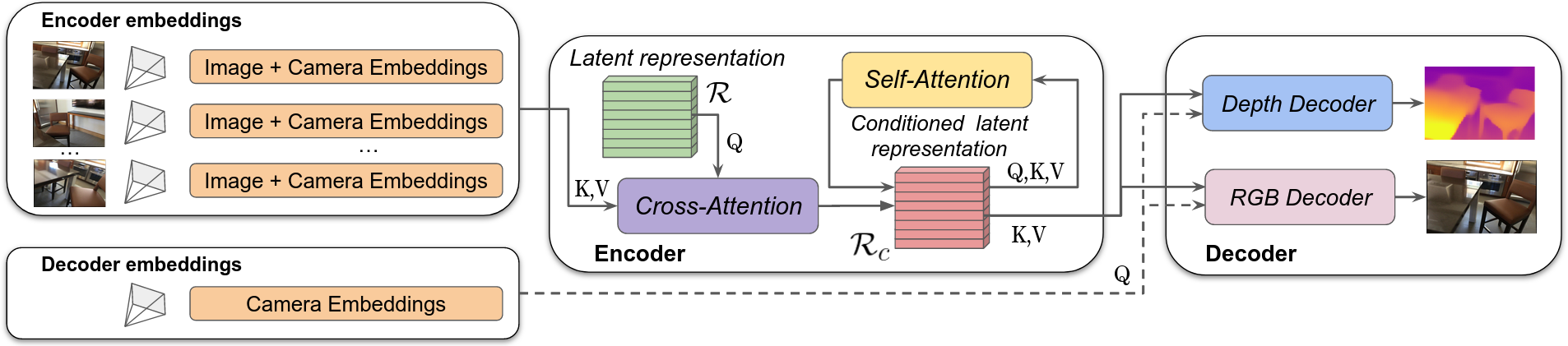}
}
\\
\subfloat[Image embeddings.]{
\label{fig:rgb_embeddings}
\includegraphics[width=0.3\textwidth]{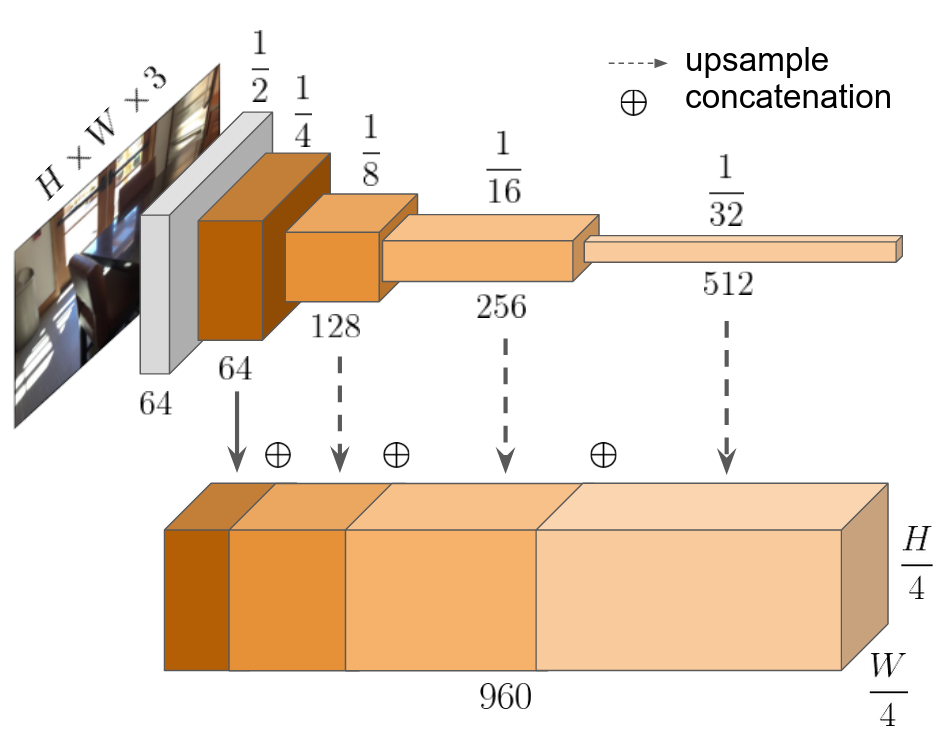}
}
\hspace{2mm}
\subfloat[Camera embeddings.]{
\label{fig:cam_embeddings}
\includegraphics[width=0.45\textwidth]{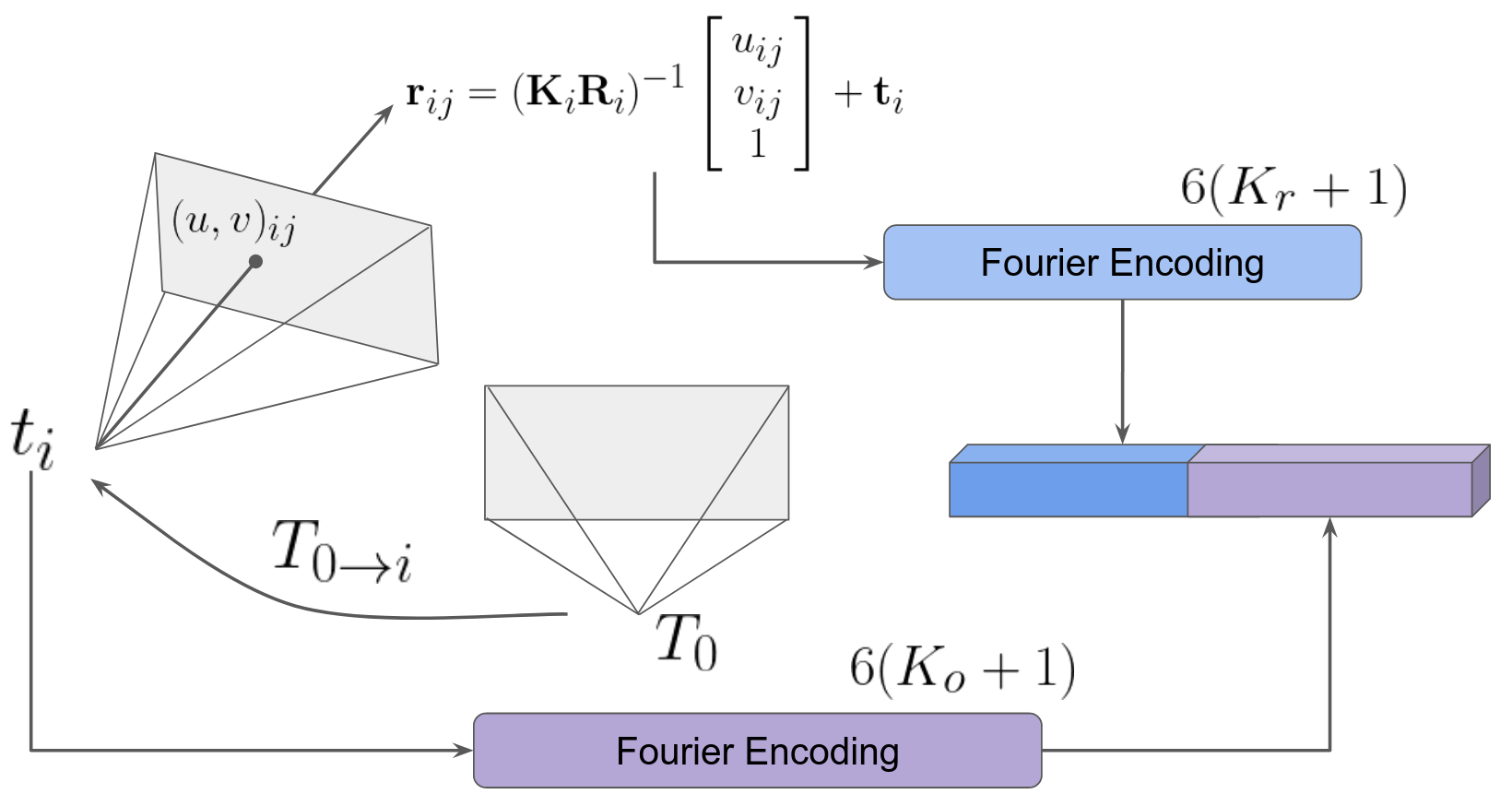}
}
\caption{
\textbf{Overview of our proposed \MethodAcronym architecture}, and the embeddings used to encode and decode information for depth and view synthesis. 
}
\label{fig:encodings}
\vspace{-6mm}
\end{figure}

\vspace{-2mm}
Our proposed \MethodAcronym architecture (Figure~\ref{fig:diagram}) is designed with flexibility in mind, so data from different sources can be used as input and different output tasks can be estimated from the same latent space. Similar to Yifan et al.~\cite{yifan2021input}, we use Perceiver~IO~\cite{jaegle2021perceiver} as our general-purpose Transformer backbone. During the encoding stage, our model takes RGB images from calibrated cameras, with known intrinsics and relative poses. The architecture processes this information according to the modality into different pixel-wise embeddings that serve as input to our Perceiver~IO backbone.  This encoded information can be queried using only camera embeddings, producing estimates from arbitrary viewpoints.  

\vspace{-2mm}
\subsection{Perceiver~IO}
Perceiver~IO~\cite{jaegle2021perceiverio} is a recent extension of the Perceiver~\cite{jaegle2021perceiver} architecture. 
The Perceiver architecture alleviates one of the main weaknesses of Transformer-based methods, namely the quadratic scaling of self-attention with input size.  This is achieved by using a fixed-size $N_l \times C_l$ latent representation $\mathcal{R}$, and learning to project high-dimensional $N_e \times C_e$ encoding embeddings onto this latent space using cross-attention layers. Self-attention is performed in the lower-dimensional latent space, producing a \emph{conditioned latent representation} $\mathcal{R}_{c}$ that can be queried using $N_d \times C_d$ decoding embeddings to generate estimates, again using cross-attention layers. 



\vspace{-2mm}
\subsection{Input-Level Embeddings}


\subsubsection{Image Embeddings (Figure \ref{fig:rgb_embeddings}).}
Input $3 \times H \times W$ images are processed using a ResNet18~\cite{he2016deep} encoder, producing a list of features maps at increasingly lower resolutions and higher dimensionality. Feature maps at $1/4$ the original resolution are concatenated with lower-resolution feature maps, after upsampling using bilinear interpolation. The resulting image embeddings are of shape $H/4 \times W/4 \times 960$, and are used in combination with the camera embeddings from each corresponding pixel (see below) to encode visual information. 
\vspace{-4mm}
\subsubsection{Camera Embeddings (Figure \ref{fig:cam_embeddings}).}
\label{sec:camera_embeddings}
These embeddings capture multi-view scene geometry (e.g., camera intrinsics and extrinsics) as additional inputs during the learning process. Let $\textbf{x}_{ij} = (u,v)$ be an image coordinate corresponding to pixel $i$ in camera $j$, with assumed known pinhole $3 \times 3$ intrinsics $\mathbf{K}_j$ and $4 \times 4$ transformation matrix $T_j= \left[
\begin{smallmatrix}
\mathbf{R}_j & \textbf{t}_j \\
\textbf{0} & 1
\end{smallmatrix}\right]$
relative to a canonical camera $T_0$. Its origin $\mathbf{o}_j$ and direction $\textbf{r}_{ij}$ are given by:
\begin{equation}
\textbf{o}_j = - \mathbf{R}_j \mathbf{t}_j 
\quad , \quad 
\textbf{r}_{ij} = \big(\mathbf{K}_j \mathbf{R}_j \big)^{-1}  
\left[u_{ij},v_{ij},1\right]^T 
+ \textbf{t}_j
\end{equation}
Note that this formulation differs from the standard convention~\cite{mildenhall2020nerf}, which does not consider the camera translation $\textbf{t}_j$ when generating viewing rays $\textbf{r}_{ij}$. We ablate this variation in Table \ref{tab:ablation}, showing that it leads to better performance for the task of depth estimation.
These two vectors are then Fourier-encoded dimension-wise to produce higher-dimensional vectors, with a mapping of:
\begin{equation}
    x \mapsto 
    \begin{bmatrix}
        x, & \sin(f_1\pi x), & \cos(f_1\pi x), & \dots, & \sin(f_K\pi x), & \cos(f_K\pi x)
    \end{bmatrix}^\top
\end{equation}
where $K$ is the number of Fourier frequencies used ($K_o$ for the origin and $K_r$ for the ray directions), equally spaced in the interval $[1,\frac{\mu}{2}]$. The resulting camera embedding is of dimensionality 
$
2 \big( 3(K_o + 1) + 3(K_r + 1) \big) = 
6 \left( K_o + K_r + 2\right)
$. During the encoding stage, camera embeddings are produced per-pixel assuming a camera with $1/4$ the original input resolution, resulting in a total of $\frac{HW}{16}$ vectors. During the decoding stage, embeddings from cameras with arbitrary calibration (i.e., intrinsics and extrinsics) can be queried to produce virtual estimates.  



\vspace{-2mm}
\subsection{Geometric 3D Augmentations}
\label{sec:augmentations}
Data augmentation is a core component of deep learning pipelines~\cite{shorten2019survey} that improves model robustness by applying transformations to the training data consistent with the data distribution in order to introduce desired equivariant properties. 
In computer vision and depth estimation in particular, standard data augmentation techniques are usually constrained to the 2D space and include color jittering, flipping, rotation, cropping, and resizing~\cite{monodepth2,yifan2021input}. 
Recent works have started looking into 3D augmentations~\cite{sajjadi2021scene} to improve robustness to errors in scene geometry in terms of camera localization (i.e., extrinsics) and parameters (i.e., intrinsics).
Conversely, we are interested in \emph{encoding} scene geometry at the input-level, so our architecture can learn a multi-view-consistent geometric latent scene representation. Therefore, in this section we propose a series of 3D augmentations to increase the number and diversity of training views while maintaining the spatial relationship between cameras, thus enforcing desired equivariant properties within this setting.


\begin{figure}[t!]
\centering
\subfloat[Virtual Camera Projection.]{
\label{fig:virtual_projection}
\includegraphics[width=0.48\textwidth]{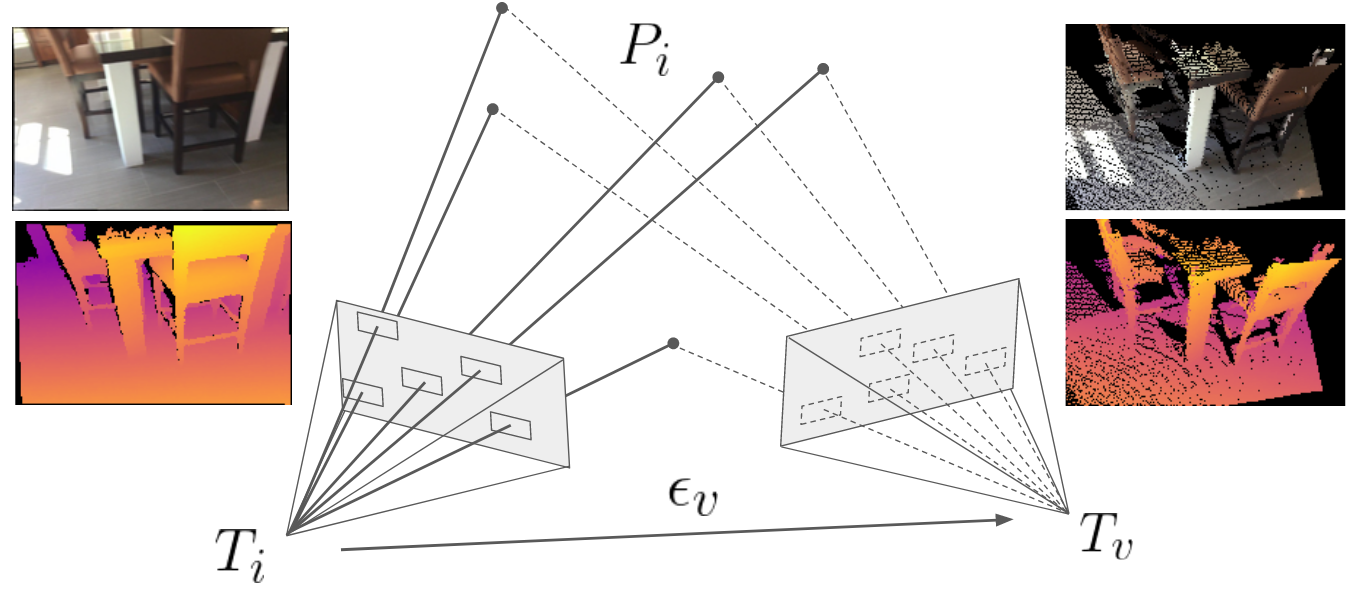}
}
\hspace{0.8mm}
\subfloat[Canonical Jittering.]{
\label{fig:canonical_jittering}
\includegraphics[width=0.45\textwidth]{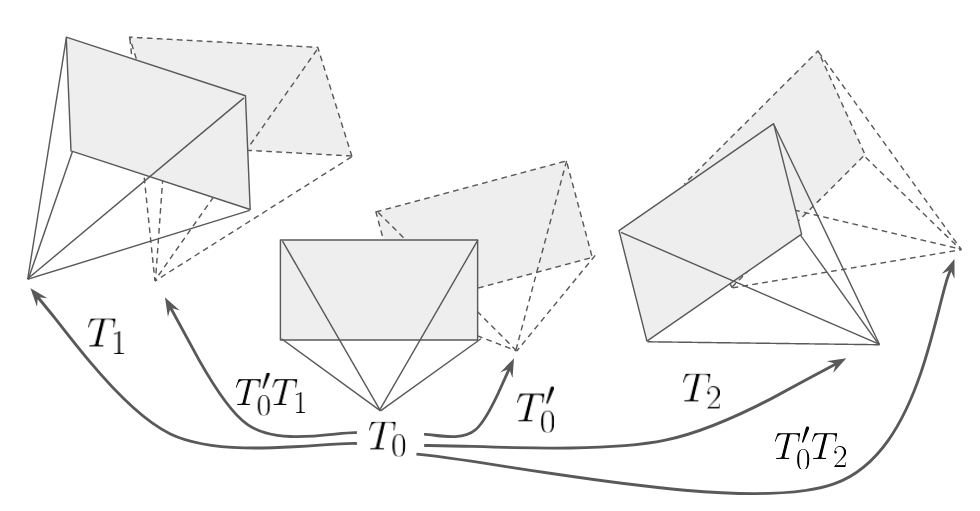}
}
\caption{
\textbf{Geometric augmentations}. 
(a) Information from camera $i$ is projected onto a virtual camera at $T_v$, creating additional supervision from other viewpoints. 
(b) Noise $T_0'$ is introduced to the canonical camera at $T_0$, and then propagated to other cameras to preserve relative scene geometry. 
}
\label{fig:augmentations}
\vspace{-3mm}
\end{figure}
\vspace{-3mm}
\subsubsection{Virtual Camera Projection.}
One of the key properties of our architecture is that it enables querying from arbitrary viewpoints, since only camera information (viewing rays) is required at the decoding stage. When generating predictions from these novel viewpoints, the network creates \emph{virtual} information consistent with the implicit structure of the learned latent scene representation, conditioned on information from the encoded views. We evaluate this capability in Section~\ref{sec:depth_synthesis}, showing superior performance relative to the explicit projection of information from encoded views. 
Here, we propose to leverage this property at training time as well, generating additional supervision in the form of \emph{virtual cameras} with corresponding ground-truth RGB images and depth maps obtained by projecting available information onto these new viewpoints (Figure~\ref{fig:virtual_projection}). This novel augmentation technique forces the learned latent scene representation to be viewpoint-independent. Experiments show that this approach provides benefits in both the (a) stereo setting, with only two viewpoints; and (b) video setting, with similar viewpoints and a dominant camera direction.

From a practical perspective, virtual cameras are generated by adding translation noise $\bm{\epsilon}_v = [\epsilon_x,\epsilon_y,\epsilon_z]_v \sim \mathcal{N}(0,\sigma_v)$ to the pose of a camera $i$. The viewing angle is set to point towards the center $\textbf{c}_i$ of the pointcloud $P_i$ generated by unprojecting information from the selected camera, which is also perturbed by $\bm{\epsilon}_c = [\epsilon_x,\epsilon_y,\epsilon_z]_c \sim \mathcal{N}(0,\sigma_v)$. When generating ground-truth information, we project the combined pointcloud from all available cameras onto these new viewpoints as a way to preserve full scene geometry. Furthermore, because the resulting RGB image and depth map will be sparse, we can improve efficiency by only querying at these specific locations. 


\vspace{-3mm}
\subsubsection{Canonical Jittering.}
When operating in a multi-camera setting, it is standard practice to select one camera to be the \emph{reference} camera, and position all other cameras relative to it~\cite{im2019dpsnet}.  One drawback of this convention is that one camera will always be at the same location (the origin of its own coordinate system) and will therefore produce the same camera embeddings, leading to overfitting. Intuitively, scene geometry should be invariant to the translation and rotation of the entire sensor suite. To enforce this property on our learned latent scene representation, we propose to inject some amount of noise to the canonical pose itself, so it is not located at the origin of the coordinate system.  
Note that this is different from methods that inject per-camera noise~\cite{novotny2017learning} with the goal of increasing robustness to localization errors. We only inject noise \emph{once}, on the canonical camera, and propagate it to other cameras, so relative scene geometry is preserved within a translation and rotation offset (Figure~\ref{fig:canonical_jittering}). However, this offset is reflected on the input-level embeddings produced by each camera, and thus forces the latent representation to be invariant to these transformations. 

In order to perform canonical jittering, we randomly sample translation $\bm{\epsilon}_t = [\epsilon_x, \epsilon_y, \epsilon_z]^\top \sim \mathcal{N}(0,\sigma_t)$ and rotation \mbox{$\bm{\epsilon}_r  = [\epsilon_\phi, \epsilon_\theta, \epsilon_\psi]^\top \sim \mathcal{N}(0,\sigma_r)$} 
errors from zero-mean normal distributions with pre-determined standard deviations.  Represented as Euler angles, we convert each set of rotation errors to a $3 \times 3$ rotation matrix $\mathbf{R}_r$. We then use the rotation matrix and translation error to create a jittered canonical transformation matrix $T_0' = \left[
\begin{smallmatrix}
\mathbf{R}_r & \bm{\epsilon}_t \\
\textbf{0} & 1
\end{smallmatrix} \right]$ that is then propagated to all other $N$ cameras, such that $T_i' = T_0' \cdot T_i$,  $\forall i \in \{1, \dots, N-1\}$.



\vspace{-3mm}
\subsubsection{Canonical Randomization.}
As an extension to canonical jittering, we also introduce canonical randomization to encourage generalization to different relative camera configurations, while still preserving scene geometry. Assuming $N$ cameras, we randomly select $o \in \{0,\dots,N-1\}$ as the canonical index. We then compute the relative transformation matrix $T_i'$ given world-frame transformation matrix $T_i$ as $T_i' = T_i \cdot T_o^{-1}$ $\forall i \in \{0,\dots,N-1\}$. 

\vspace{-2mm}
\subsection{Decoders}

We use task-specific decoders, each consisting of one cross-attention layer between the $N_d \times C_d$ queries and the $N_l \times C_l$ conditioned latent representation $\mathcal{R}_c$, followed by a linear layer that creates an output of size $N_d \times C_o$, and a sigmoid activation function $\sigma(x)=\frac{1}{1 + e^{-x}}$ to produce values in the interval $[0,1]$. 
We set $C_o^d = 1$ for depth estimation and $C_o^s = 3$ for view synthesis. Depth estimates are scaled to lie within the range $[d_\text{min}, d_\text{max}]$. Note that other decoders can be incorporated with \MethodAcronym without any modification to the underlying architecture, enabling the generation of multi-task estimates from arbitrary viewpoints. 


\vspace{-2mm}
\subsection{Losses}

We use an L1-log loss $\mathcal{L}_{d} = \lVert \log(d_{ij}) - \log(\hat{d}_{ij})\rVert_1$ to supervise depth estimation, where $\hat{d}_{ij}$ and $d_{ij}$ are depth estimates and ground truth, respectively, for pixel $j$ at camera $i$. For view synthesis, we use an L2 loss $\mathcal{L}_{s} = \lVert \textbf{p}_{ij} - \hat{\textbf{p}}_{ij} \rVert^2$, where $\hat{\textbf{p}}_{ij}$ and $\textbf{p}_{ij}$ are RGB estimates and ground truth, respectively, for pixel $j$ at camera $i$. We use a weight coefficient $\lambda_{s}$ to balance these two losses, and another $\lambda_{v}$ to balance losses from available and virtual cameras. The final loss is of the form:
\begin{equation}
\label{eq:loss}
\mathcal{L} = \mathcal{L}_{d} + \lambda_{s} \mathcal{L}_{s} + \lambda_{v}\big(\mathcal{L}_{d,v} + \lambda_{s}(\mathcal{L}_{s,v})\big)
\end{equation}
Note that because our architecture enables querying at specific image coordinates, we can improve efficiency at training time by not computing estimates for pixels without ground truth (e.g., sparse depth maps or virtual cameras). 




\section{Experiments}



\subsection{Datasets}

\subsubsection{ScanNet~\cite{dai2017scannet}}
We evaluate our \MethodAcronym for both \emph{stereo} and \emph{video} depth estimation using ScanNet,  an RGB-D video dataset that contains $2.5$ million views from around $1500$ scenes.
For the stereo experiments, we follow the same setting as Kusupati et al.~\cite{kusupati2020normal}: we downsample scenes by a factor of $20$ and use a custom split to create stereo pairs, resulting in $94212$ training and $7517$ test samples. 
For the video experiments, we follow the evaluation protocol of Teed et al.~\cite{deepv2d}, with a total of $1405$ scenes for training.
For the test set, we use a custom split to select $2000$ samples from $90$ scenes not covered in the training set. Each training sample includes a target frame and a context of $[-3,3]$ frames with stride $3$. Each test sample includes a pair of frames, with a context of $[-3,3]$ relative to the first frame of the pair with stride $3$.


\vspace{-5mm}
\subsubsection{7-Scenes~\cite{shotton2013scene}}
We also evaluate on the test split of 7-Scenes to measure zero-shot cross-dataset performance. Collected using KinectFusion~\cite{KinectFusion}, the dataset consists of $640 \times 480$ images in $7$ settings, with a variable number of scenes in each setting.  There are $500$--$1000$ images in each scene.  We follow the evaluation protocol of Sun et al.~\cite{Sun_2021_CVPR}, median-scaling predictions using ground-truth information before evaluation. 

\vspace{-2mm}
\subsection{Stereo Depth Estimation}
\label{sec:stereo_depth_estimation}
\begin{figure}[t!]
\renewcommand{\arraystretch}{0.95}
\centering
\subfloat[Depth estimation results.]{
\label{fig:stereo_results}
\raisebox{13mm}{
    \begin{tabular}{l|ccc}
        \toprule
        \textbf{Method} &
        \small{Abs.Rel}$\downarrow$ &
        RMSE$\downarrow$ &
        $\delta_{1.25}$$\uparrow$ \\
        \toprule
        DPSNet~\cite{im2019dpsnet} & 0.126 & 0.314 & ---  \\
        NAS~\cite{kusupati2020normal} & 0.107& 0.281 & ---  \\
        IIB~\cite{yifan2021input} & 0.116  & 0.281 & 0.908  \\
        \midrule
        \textbf{\MethodAcronym} ($128 \times 192$) & 0.093  & 0.246 & 0.911 \\
        \textbf{\MethodAcronym} ($240 \times 320$) & \textbf{0.089} & \textbf{0.232} & \textbf{0.915} \\
        \bottomrule
    \end{tabular}
}
\label{tab:depth_scannet_stereo}
}
\subfloat[Virtual depth estimation results.]{
\label{fig:virtual_results}
    \includegraphics[width=0.40\textwidth, height=2.8cm]{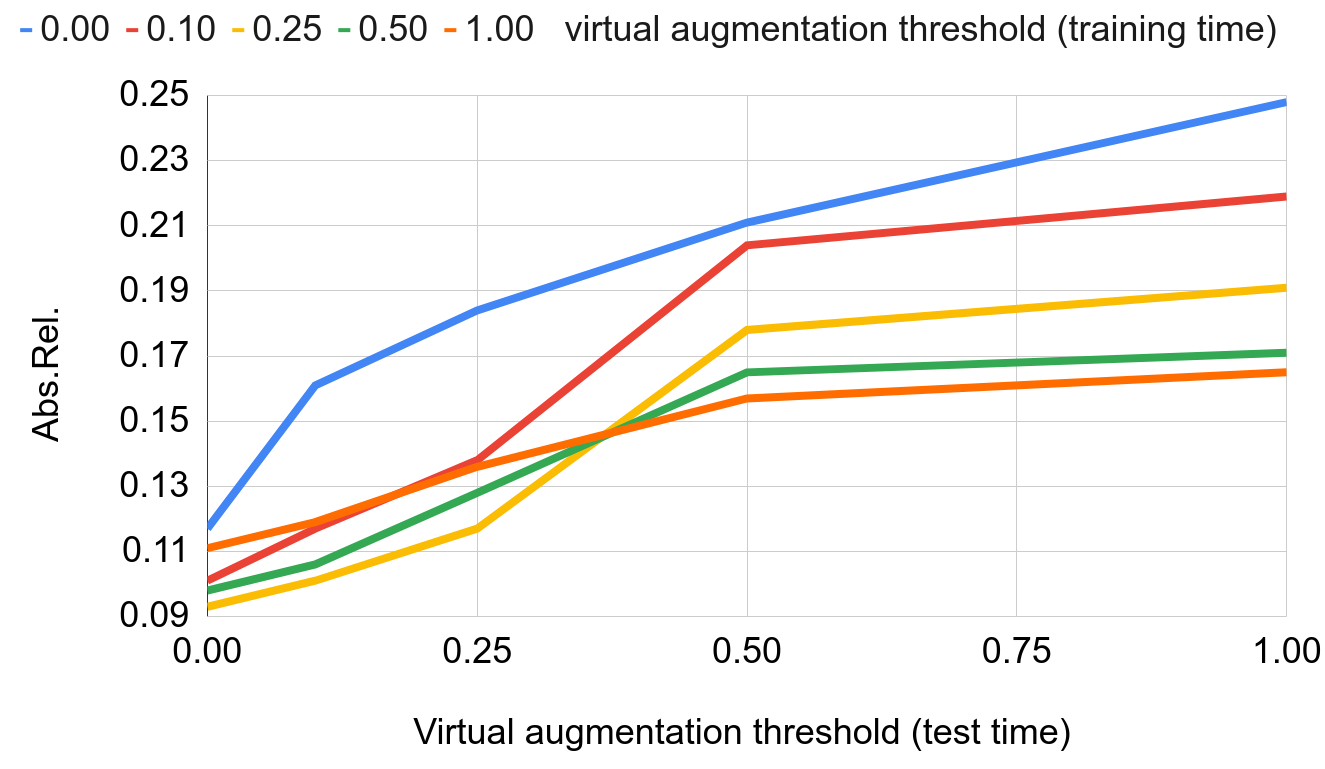}
    \label{fig:virtual_error}
}
\vspace{-2mm}
\caption{\textbf{Depth estimation results on ScanNet-Stereo}. (a) We outperform contemporary methods by a large margin. (b) Depth estimation results on virtual cameras using different values for $\sigma_v$ at training and test time. 
}
\vspace{-1mm}
\label{fig:depth_scannet_stereo}
\end{figure}

\begin{figure}[t!]
\centering
\rotatebox{90}{\hspace{6mm}\tiny{GT}} \!
\subfloat{\includegraphics[width=0.22\textwidth,height=1.6cm,trim={20cm 0 0 0},clip]{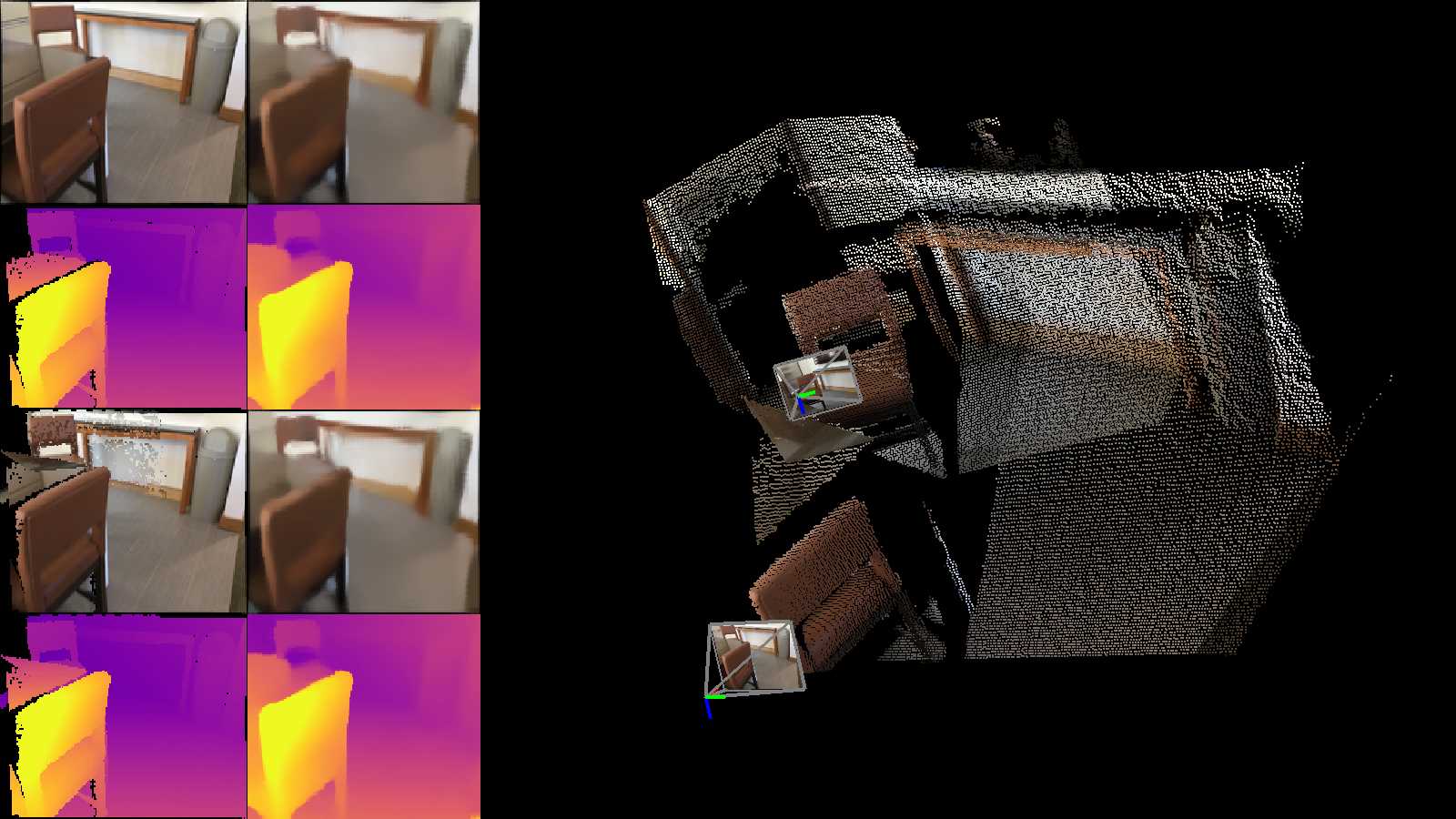}} \hspace{1mm}
\subfloat{\includegraphics[width=0.22\textwidth,height=1.6cm,trim={20cm 0 0 0},clip]{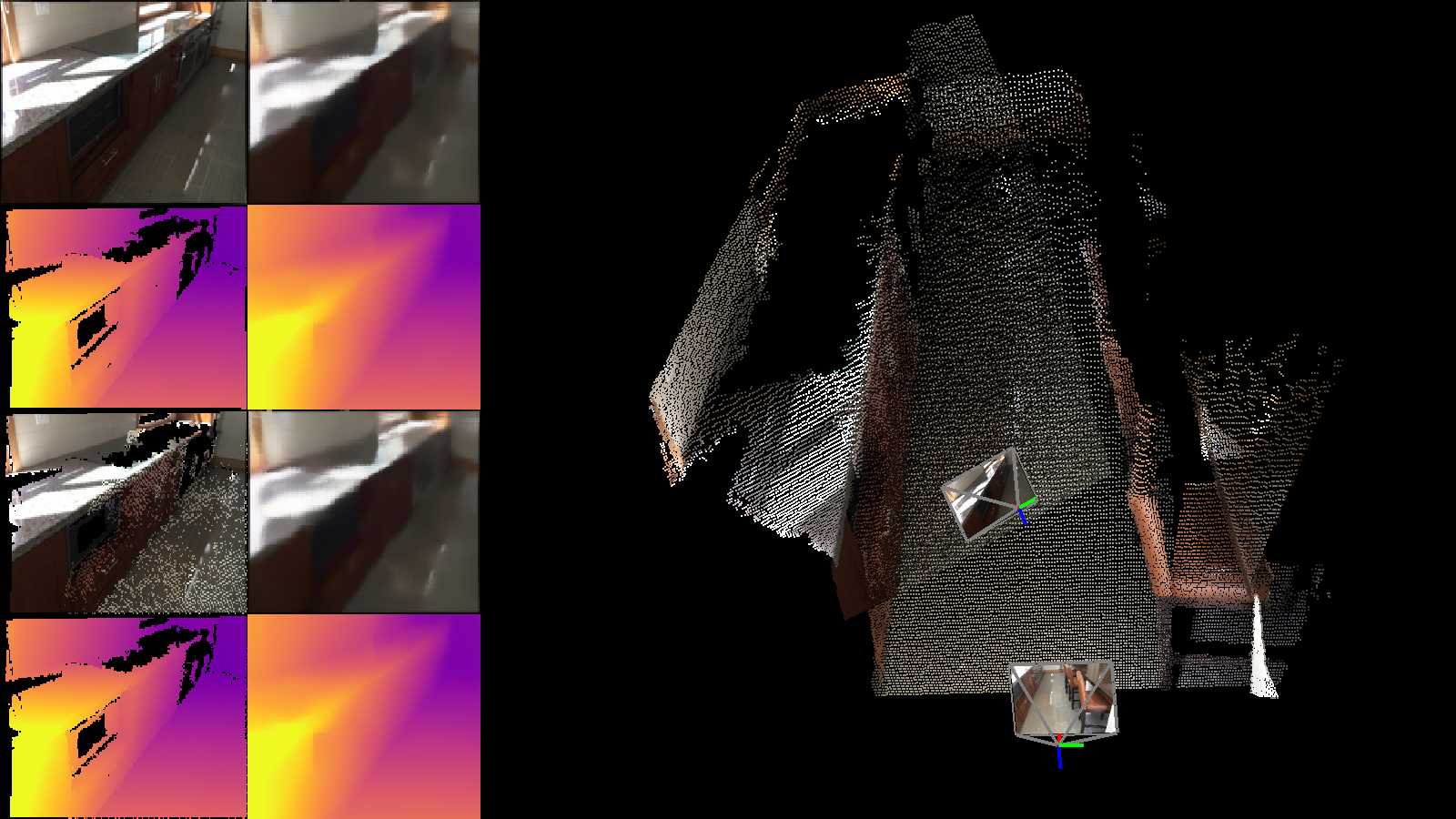}} \hspace{1mm}
\subfloat{\includegraphics[width=0.22\textwidth,height=1.6cm,trim={20cm 0 0 0},clip]{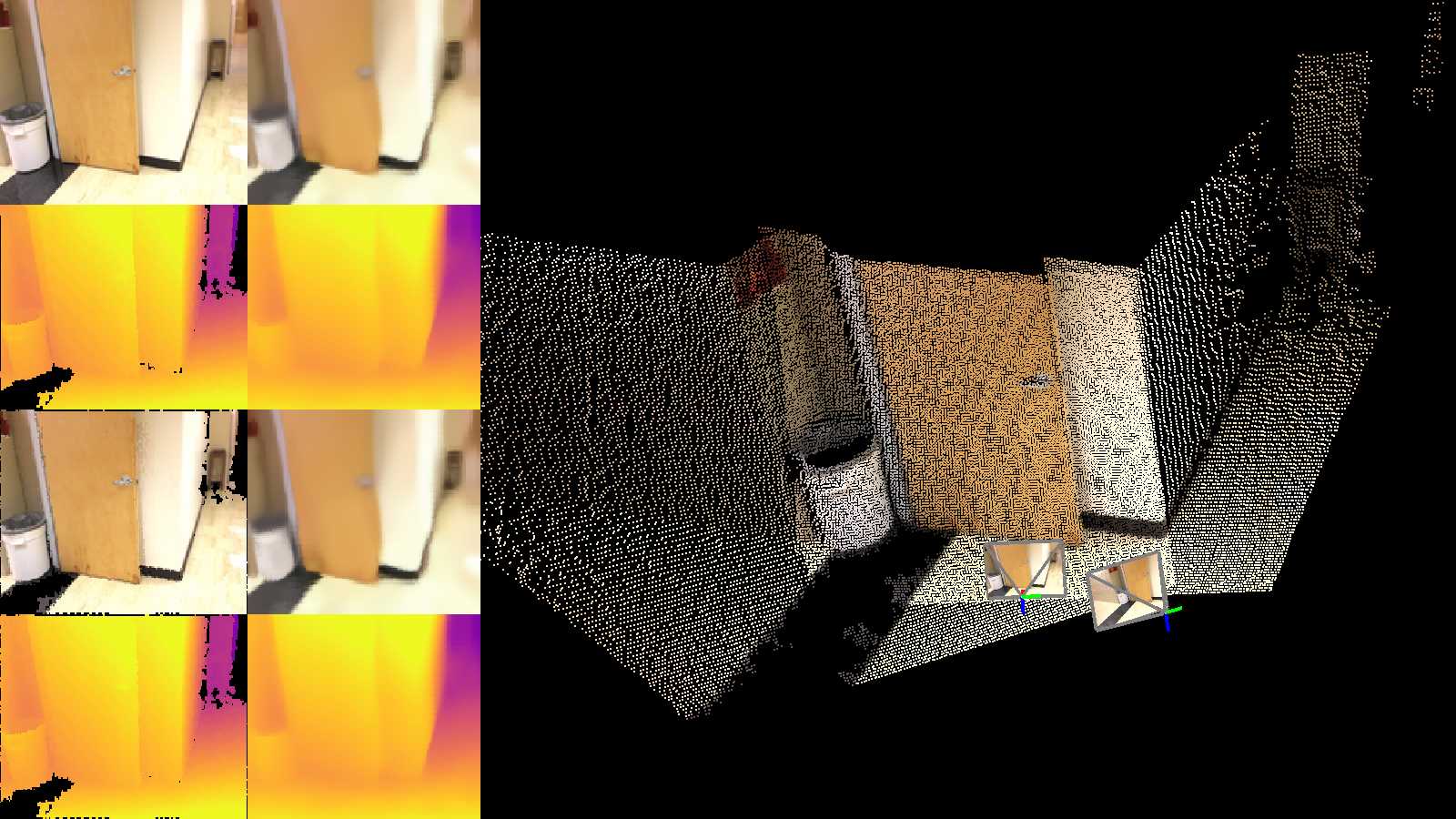}} \hspace{1mm}
\subfloat{\includegraphics[width=0.22\textwidth,height=1.6cm,trim={20cm 0 0 0},clip]{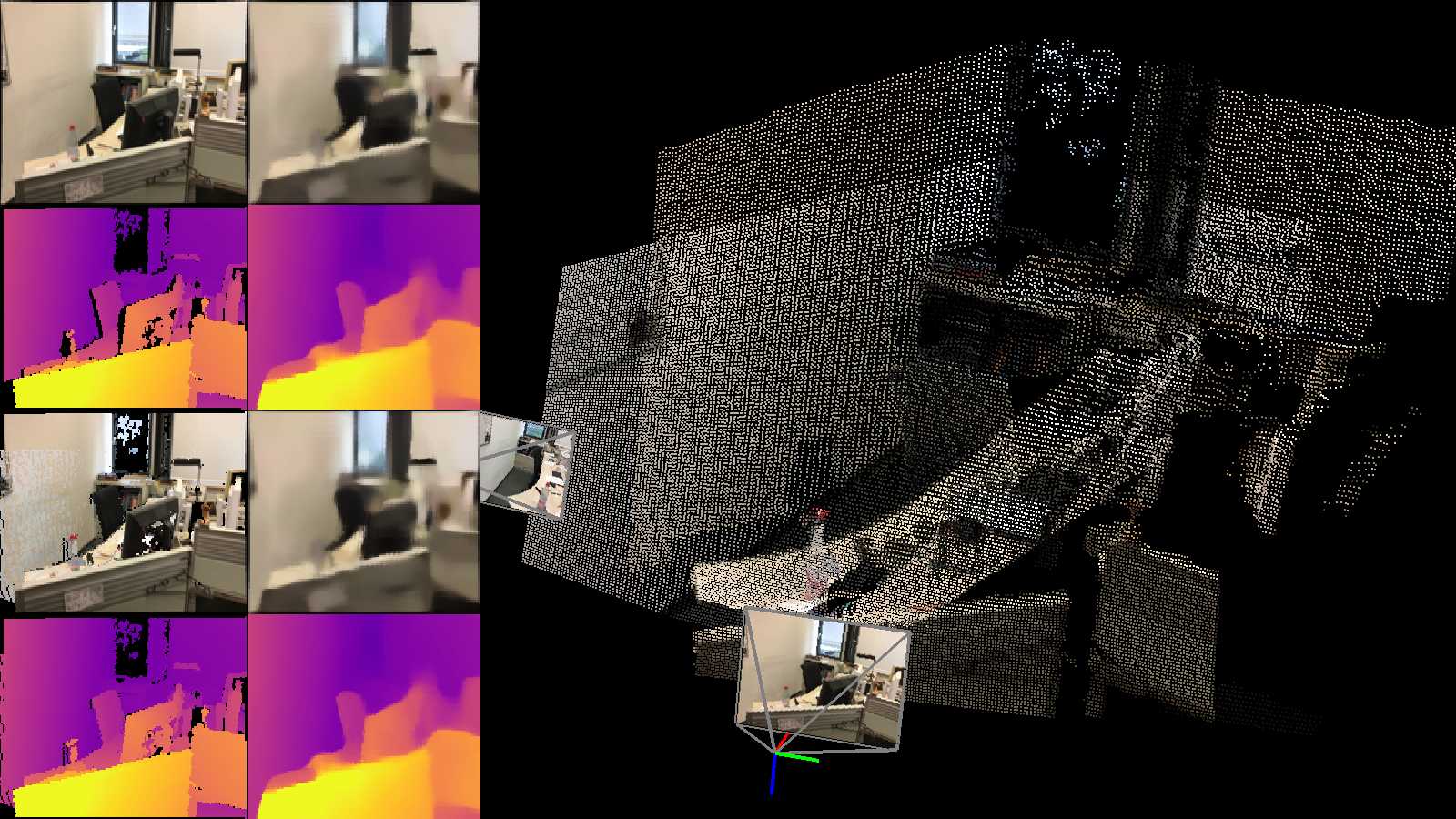}}
\\ 
\rotatebox{90}{\hspace{4mm}\tiny{\MethodAcronym}} \!
\subfloat{\includegraphics[width=0.22\textwidth,height=1.6cm,trim={20cm 0 0 0},clip]{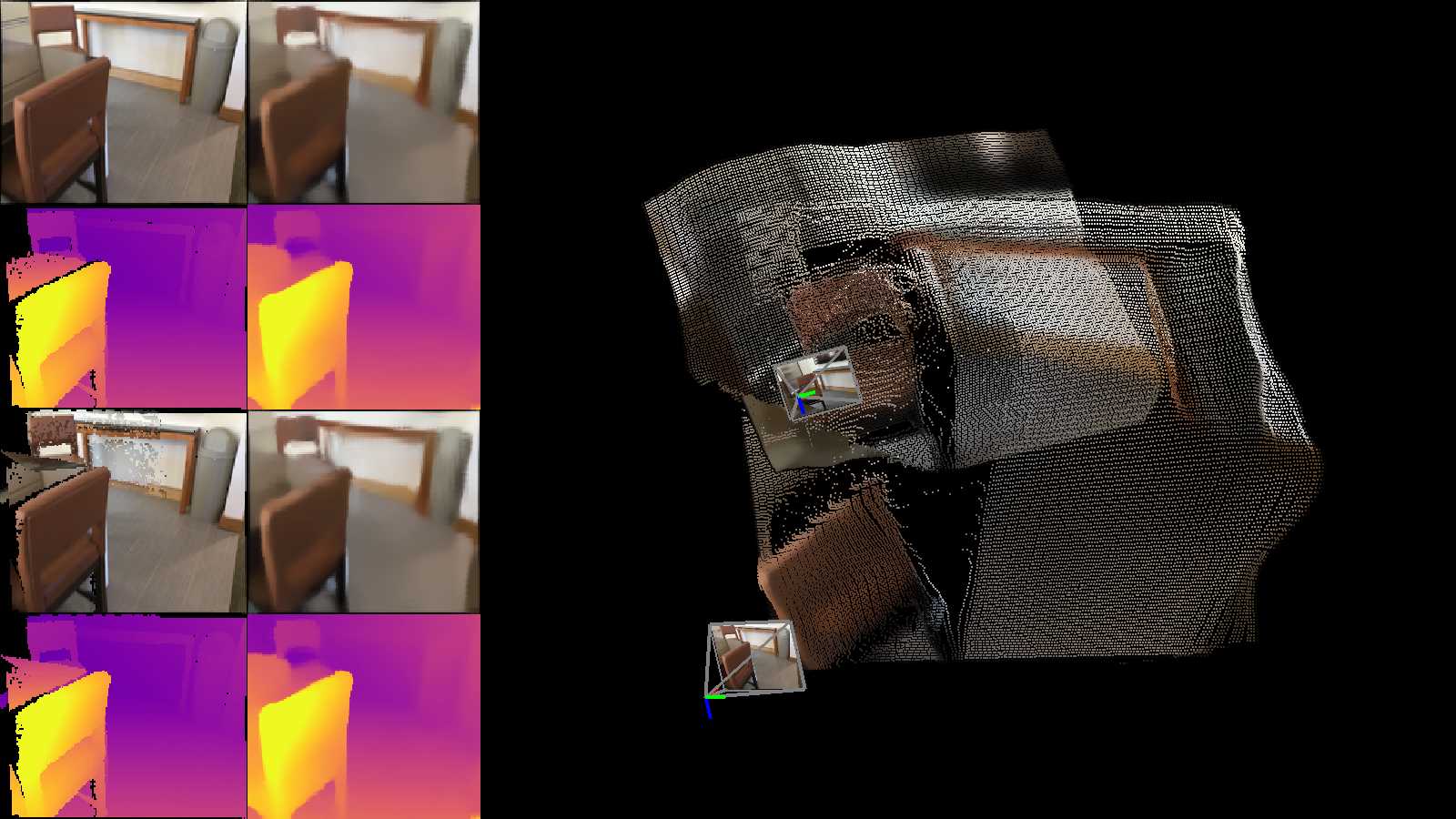}} \hspace{1mm}
\subfloat{\includegraphics[width=0.22\textwidth,height=1.6cm,trim={20cm 0 0 0},clip]{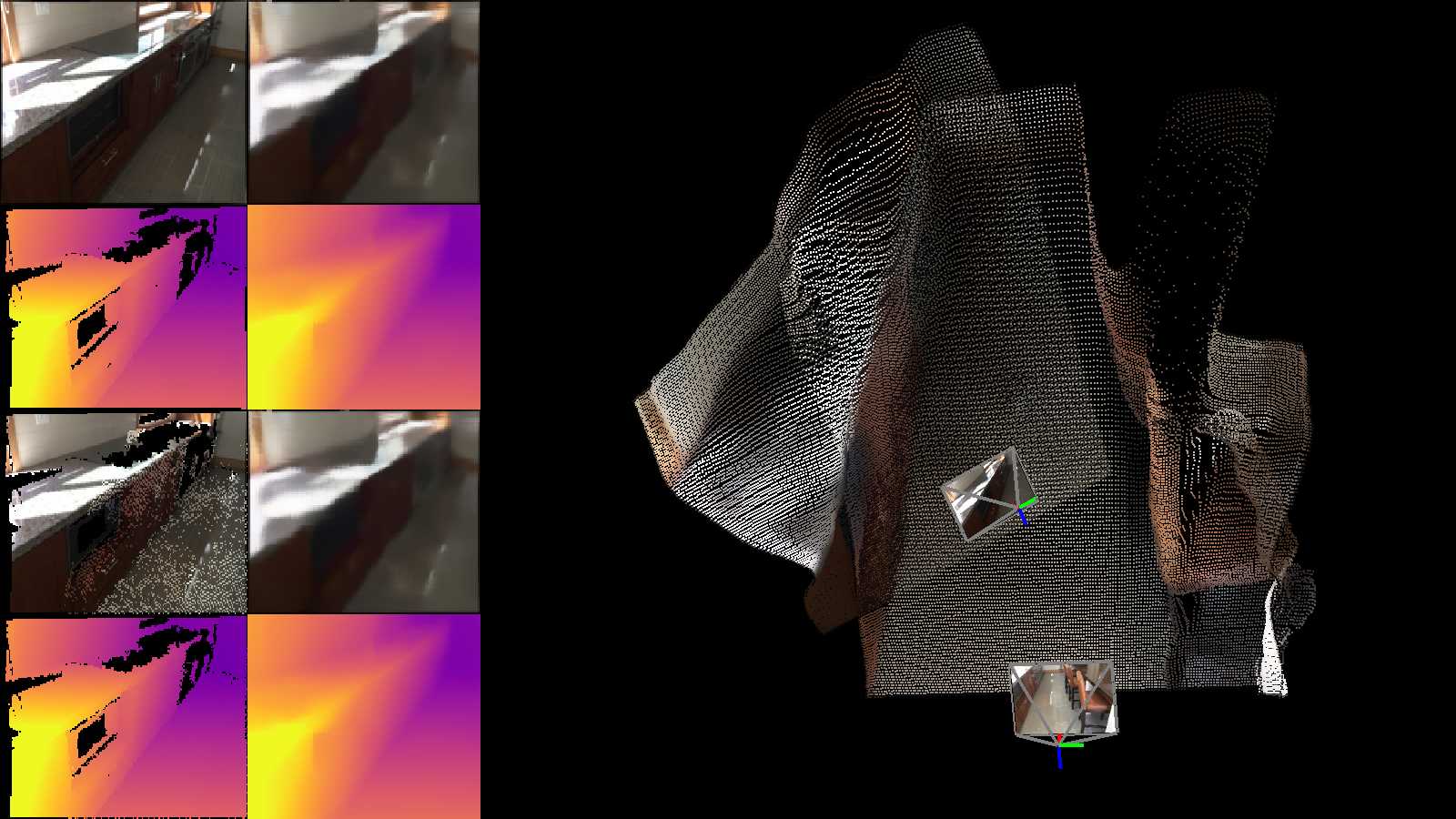}} \hspace{1mm}
\subfloat{\includegraphics[width=0.22\textwidth,height=1.6cm,trim={20cm 0 0 0},clip]{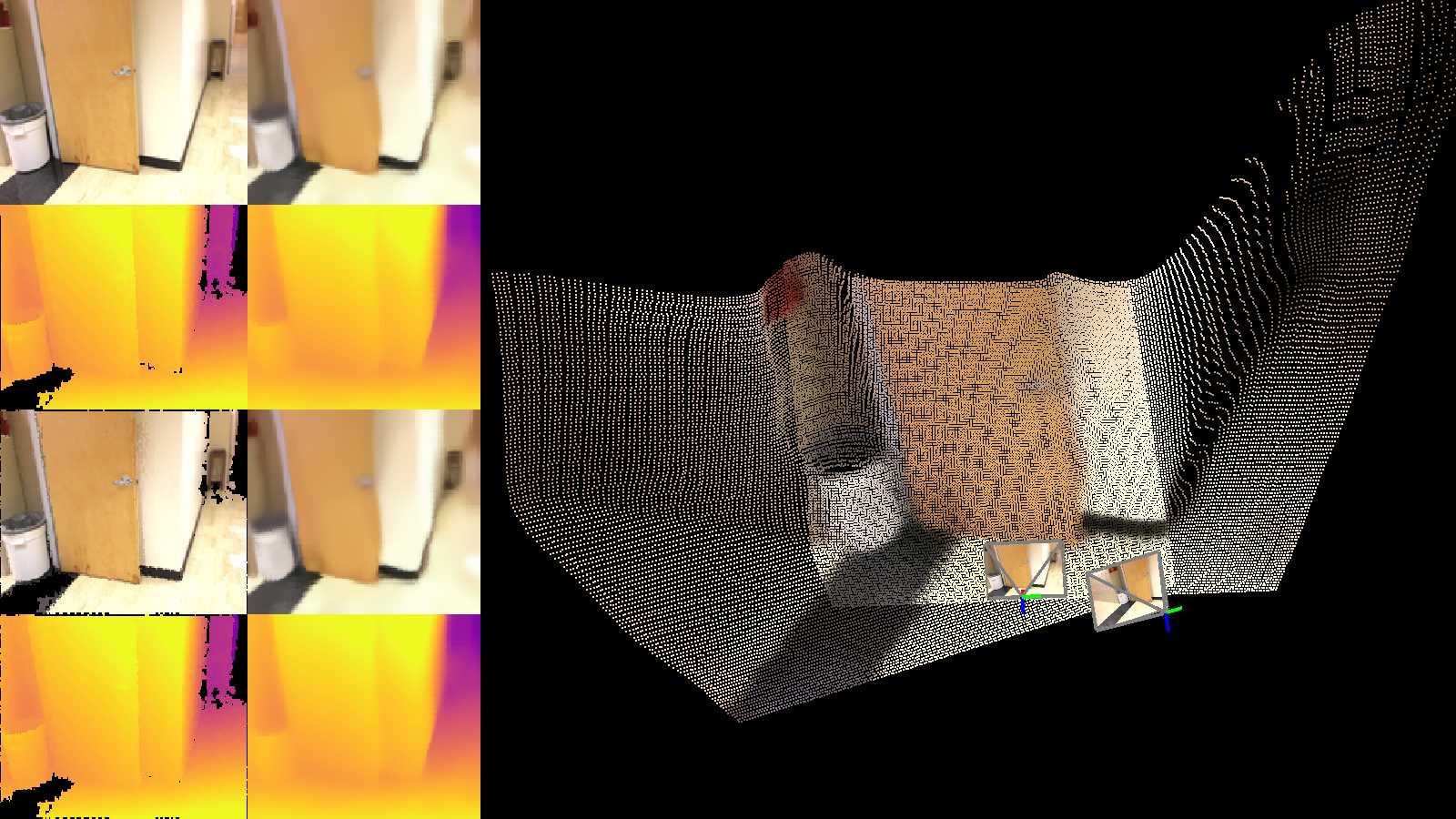}} \hspace{1mm}
\subfloat{\includegraphics[width=0.22\textwidth,height=1.6cm,trim={20cm 0 0 0},clip]{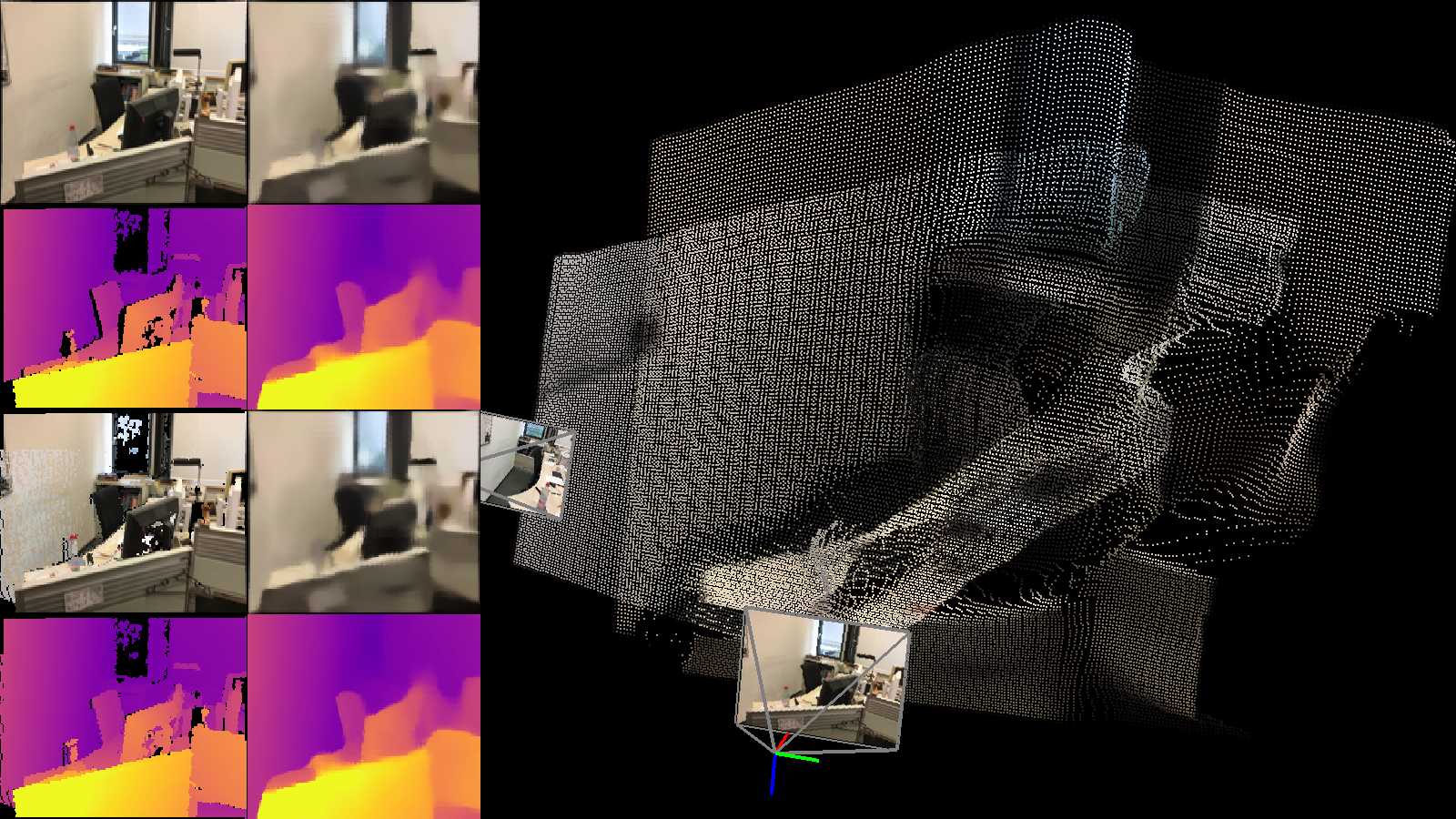}}
\caption{\textbf{Reconstructed two-view pointclouds}, from ScanNet-Stereo. \MethodAcronym pointclouds are generated using both depth maps and RGB images queried from our learned latent representation.}
\label{fig:pointclouds}
\end{figure}

To test the benefits our proposed geometric 3D augmentation procedures over the IIB~\cite{yifan2021input} baseline, we first evaluate \MethodAcronym on the task of stereo depth estimation. Here, because each sample provides minimal information about the scene (i.e., only two frames), the introduction of additional virtual supervision should have the largest effect. We report our results in Figure~\ref{fig:depth_scannet_stereo}a and visualize examples of reconstructed pointclouds in Figure~\ref{fig:pointclouds}. \MethodAcronym significantly outperforms other methods on this dataset, including IIB.  Our virtual view augmentations lead to a large ($20\%$) relative improvement, showing that \MethodAcronym benefits from a scene representation that encourages multi-view consistency.

\begin{table*}[t!]
\renewcommand{\arraystretch}{0.92}
\centering
{
\small
\setlength{\tabcolsep}{0.3em}
\begin{tabular}{c|l|ccc|ccc}
\toprule
\multicolumn{2}{c|}{\multirow{2}{*}{\textbf{Variation}}} &
\multicolumn{3}{c|}{Lower is better $\downarrow$} &
\multicolumn{3}{c}{Higher is better $\uparrow$}
 \\
& & 
Abs.\ Rel &
Sq.\ Rel &
RMSE &
$\delta_{1.25}$ &
$\delta_{1.25^2}$ &
$\delta_{1.25^3}$ \\
\toprule
1 & Depth-Only & 0.098 & 0.046 & 0.257 & 0.902 & 0.972 & 0.990 \\
\midrule
2 & w/ Conv. RGB encoder~\cite{jaegle2021perceiverio} & 0.114 & 0.058 & 0.294 & 0.866 & 0.961 & 0.982  \\
3 & w/ 64-dim R18 RGB encoder & 0.104 & 0.049 & 0.270 & 0.883 & 0.966 & 0.985  \\
\midrule
4 & w/o camera information & 0.229 & 0.157 & 0.473 & 0.661 & 0.874 & 0.955  \\
5 & w/o global rays encoding & 0.097 & 0.047 & 0.261 & 0.897 & 0.962 & 0.988  \\
6 & w/ equal loss weights & 0.095 & 0.047 & 0.259 & 0.908 & 0.968 & 0.990  \\
7 & w/ epipolar cues~\cite{yifan2021input} & 0.094 & 0.048 & 0.254 & 0.905 & 0.972 & 0.990 \\
\midrule
8 & w/o Augmentations & 0.117 & 0.060 & 0.291 & 0.870 & 0.959 & 0.981  \\
9 & w/o Virtual Cameras & 0.104 & 0.058 & 0.268 & 0.891 & 0.965 & 0.986  \\
10 & w/o Canonical Jittering & 0.099 & 0.046 & 0.261 & 0.897 & 0.970 & 0.988  \\
11 & w/o Canonical Randomization & 0.096 & 0.044 & 0.253 & 0.905 & 0.971 & 0.989 \\
\midrule
& \textbf{\MethodAcronym} & \textbf{0.093} & \textbf{0.042} & \textbf{0.246} & \textbf{0.911} & \textbf{0.974} & \textbf{0.991} \\
\bottomrule
\end{tabular}
}
\caption{\textbf{Ablation study for ScanNet-Stereo}, using different variations. } 
\label{tab:ablation}
\vspace{-8mm}
\end{table*}

\vspace{-4mm}
\subsection{Ablation Study}
\label{sec:ablation}

We perform a detailed ablation study to evaluate the effectiveness of each component in our proposed architecture, with results shown in Table~\ref{tab:ablation}. Firstly, we evaluate performance when (Table~\ref{tab:ablation}:1) learning only depth estimation, and see that the joint learning of view synthesis as an auxiliary task leads to significant improvements. The claim that depth estimation improves view synthesis has been noted before~\cite{deng2021depth,nerfingmvs}, and is attributed to the well-known fact that multi-view consistency facilitates the generation of images from novel viewpoints.  However, our experiments also show the inverse: view synthesis improves depth estimation. Our hypothesis is that appearance is required to learn multi-view consistency since it enables visual correlation across frames. By introducing view synthesis as an additional task, we are also encoding appearance-based information into our latent representation. This leads to improvements in depth estimation even though no explicit feature matching is performed at an architecture or loss level.

\begin{figure}[t!]
\centering
\rotatebox{90}{\hspace{2.5mm} \tiny{GT}} \!
\subfloat{\includegraphics[width=0.13\textwidth,height=1.1cm]{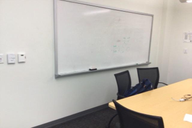}}
\hspace{0.05mm}
\subfloat{\includegraphics[width=0.13\textwidth,height=1.1cm]{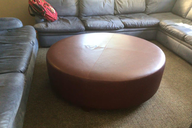}}
\hspace{0.05mm}
\subfloat{\includegraphics[width=0.13\textwidth,height=1.1cm]{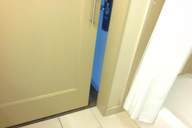}}
\hspace{0.05mm}
\subfloat{\includegraphics[width=0.13\textwidth,height=1.1cm]{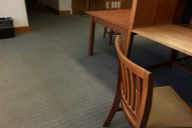}}
\hspace{0.05mm}
\subfloat{\includegraphics[width=0.13\textwidth,height=1.1cm]{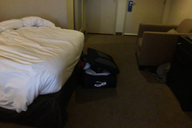}}
\hspace{0.05mm}
\subfloat{\includegraphics[width=0.13\textwidth,height=1.1cm]{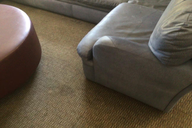}}
\hspace{0.05mm}
\subfloat{\includegraphics[width=0.13\textwidth,height=1.1cm]{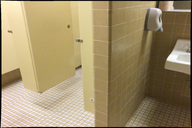}}
\\ \vspace{-1mm}
\rotatebox{90}{\hspace{0.5mm} \tiny{DeFiNe}} \!
\subfloat{\includegraphics[width=0.13\textwidth,height=1.1cm]{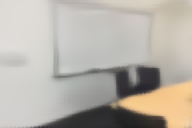}}
\hspace{0.05mm}
\subfloat{\includegraphics[width=0.13\textwidth,height=1.1cm]{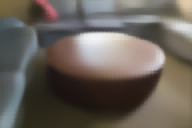}}
\hspace{0.05mm}
\subfloat{\includegraphics[width=0.13\textwidth,height=1.1cm]{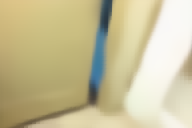}}
\hspace{0.05mm}
\subfloat{\includegraphics[width=0.13\textwidth,height=1.1cm]{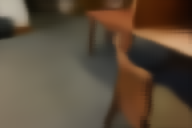}}
\hspace{0.05mm}
\subfloat{\includegraphics[width=0.13\textwidth,height=1.1cm]{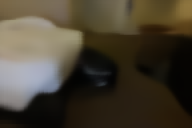}}
\hspace{0.05mm}
\subfloat{\includegraphics[width=0.13\textwidth,height=1.1cm]{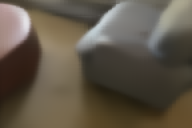}}
\hspace{0.05mm}
\subfloat{\includegraphics[width=0.13\textwidth,height=1.1cm]{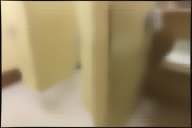}}
\\ \vspace{-1mm}
\rotatebox{90}{\hspace{2.5mm} \tiny{GT}} \!
\subfloat{\includegraphics[width=0.13\textwidth,height=1.1cm]{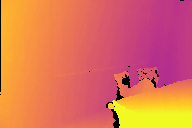}}
\hspace{0.05mm}
\subfloat{\includegraphics[width=0.13\textwidth,height=1.1cm]{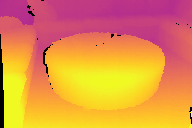}}
\hspace{0.05mm}
\subfloat{\includegraphics[width=0.13\textwidth,height=1.1cm]{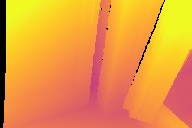}}
\hspace{0.05mm}
\subfloat{\includegraphics[width=0.13\textwidth,height=1.1cm]{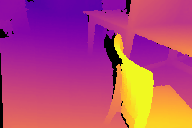}}
\hspace{0.05mm}
\subfloat{\includegraphics[width=0.13\textwidth,height=1.1cm]{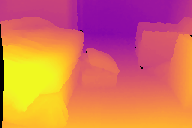}}
\hspace{0.05mm}
\subfloat{\includegraphics[width=0.13\textwidth,height=1.1cm]{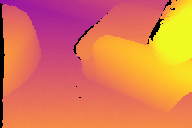}}
\hspace{0.05mm}
\subfloat{\includegraphics[width=0.13\textwidth,height=1.1cm]{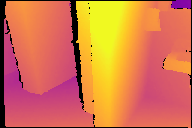}}
\\ \vspace{-1mm}
\rotatebox{90}{\hspace{0.5mm} \tiny{DeFiNe}} \!
\subfloat{\includegraphics[width=0.13\textwidth,height=1.1cm]{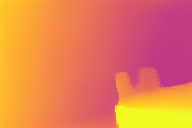}}
\hspace{0.05mm}
\subfloat{\includegraphics[width=0.13\textwidth,height=1.1cm]{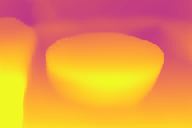}}
\hspace{0.05mm}
\subfloat{\includegraphics[width=0.13\textwidth,height=1.1cm]{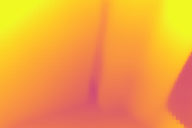}}
\hspace{0.05mm}
\subfloat{\includegraphics[width=0.13\textwidth,height=1.1cm]{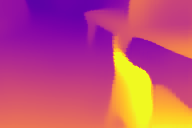}}
\hspace{0.05mm}
\subfloat{\includegraphics[width=0.13\textwidth,height=1.1cm]{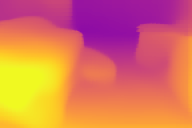}}
\hspace{0.05mm}
\subfloat{\includegraphics[width=0.13\textwidth,height=1.1cm]{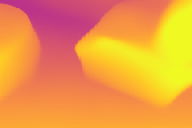}}
\hspace{0.05mm}
\subfloat{\includegraphics[width=0.13\textwidth,height=1.1cm]{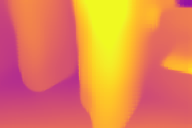}}
\caption{\textbf{Depth estimation and view synthesis results} on ScanNet. Although view synthesis is not our primary goal, it can be achieved with minimal modifications, and we show that it improves depth estimation performance.}
\label{fig:temporal_qualitative}
\vspace{-4mm}
\end{figure}

We also ablate different variations of our RGB encoder for the generation of image embeddings and show that (Table~\ref{tab:ablation}:2) our proposed multi-level feature map concatenation (Figure~\ref{fig:rgb_embeddings}) leads to the best results relative to the standard single convolutional layer, or (Table~\ref{tab:ablation}:3) using $1/4$-resolution ResNet18 64-dimensional feature maps. Similarly, we also ablate some of our design choices, namely (Table~\ref{tab:ablation}:4) the use of camera embeddings instead of positional encodings; (Table~\ref{tab:ablation}:5) global viewing rays (Section~\ref{sec:camera_embeddings}) instead of traditional relative viewing rays; (Table~\ref{tab:ablation}:6) the use of $\lambda_s=1$ in the loss calculation (Equation~\ref{eq:loss}) such that both depth and view synthesis tasks have equal weights; and (Table~\ref{tab:ablation}:7) the use of epipolar cues as additional geometric embeddings, as proposed by IIB~\cite{yifan2021input}. As expected, camera embeddings are crucial for multi-view consistency and global viewing rays improve over the standard relative formulation. Interestingly, using a smaller $\lambda_s$ degrades depth estimation performance, providing further evidence that the joint learning of view synthesis is beneficial for multi-view consistency. We did not observe meaningful improvements when incorporating the epipolar cues from IIB~\cite{yifan2021input}, indicating that \MethodAcronym is capable of directly capturing these constraints at an input-level due to the increase in viewpoint diversity. Lastly, we ablate the impact of our various proposed geometric augmentations (Section~\ref{sec:augmentations}), showing that they are key to our reported state-of-the-art performance. 

Lastly, we evaluate depth estimation from virtual cameras, using different noise levels $\sigma_v$ at test time. We also train models using different noise levels and report the results in Figure~\ref{fig:virtual_results}. From these results, we can see that the optimal virtual noise level, when evaluating at the target location, is $\sigma_v=0.25$\,m (yellow line), relative to the baseline without virtual noise (blue line). However, models trained with higher virtual noise (e.g., the orange line, with $\sigma_v=1$\,m) are more robust to larger deviations from the target location.

\vspace{-4mm}
\subsection{Video Depth Estimation}
\label{sec:video_depth_estimation}

To highlight the flexibility of our proposed architecture, we also experiment using video data from ScanNet following the training protocol of Tang et al.~\cite{tang2018ba}. We evaluate performance on both ScanNet itself, using their evaluation protocol~\cite{tang2018ba}, as well as zero-shot transfer (without fine-tuning) to the 7-Scenes dataset. 
Table~\ref{tab:depth_scannet_temporal} reports quantitative results, while Figure~\ref{fig:temporal_qualitative} provides qualitative examples.
On ScanNet, \MethodAcronym outperforms most published approaches, significantly improving over DeMoN~\cite{ummenhofer2017demon}, BA-Net~\cite{tang2018ba}, and CVD~\cite{luo2020consistent} both in terms of performance and speed. 
We are competitive with DeepV2D~\cite{deepv2d} in terms of performance, and roughly $14\times$ faster, owing to the fact that \MethodAcronym does not require bundle adjustment or any sort of test-time optimization. 
In fact, our inference time of $49$\,ms can be split into $44$\,ms for encoding and only $5$\,ms for decoding, enabling very efficient generation of depth maps after information has been encoded once. 
The only method that outperforms \MethodAcronym in terms of speed is NeuralRecon~\cite{Sun_2021_CVPR}, which uses a sophisticated TSDF integration strategy. Performance-wise, we are also competitive with NeuralRecon, improving over their reported results in one of the three  metrics (Sq.\ Rel). 

Next, we evaluate zero-shot transfer from ScanNet to 7-Scenes, which is a popular test of generalization for video depth estimation. In this setting, \MethodAcronym significantly improves over all other methods, including DeepV2D (which fails to generalize) and NeuralRecon ($\sim$$40\%$ improvement). We attribute this large gain to the highly intricate and specialized nature of these other architectures. In contrast, our method has no specialized module and instead learns a geometrically-consistent multi-view latent representation. 

In summary, we achieve competitive results on ScanNet and significantly improve the state-of-the-art for video depth generalization, as evidenced by the large gap between \MethodAcronym and the best-performing methods on 7-Scenes.

\begin{table}[t!]
\renewcommand{\arraystretch}{0.95}
\centering
    \begin{tabular}{l|ccc|c}
        \toprule
        \textbf{Method} &
        \small{Abs.\ Rel}$\downarrow$ &
        Sq.\ Rel$\downarrow$ &
        RMSE$\downarrow$ & Speed (ms)$\downarrow$
        \\
\midrule
\midrule
\multicolumn{5}{l}{\textbf{ScanNet test split} \cite{tang2018ba}} \\
\midrule
DeMoN~\cite{ummenhofer2017demon} &
0.231 & 0.520 & 0.761 & 110  \\
MiDas-v2~\cite{ranftl2020towards} & 
0.208 & 0.318 & 0.742 & - \\
BA-Net~\cite{tang2018ba} &
0.091 & 0.058 & 0.223 & 95 \\
CVD~\cite{luo2020consistent} &   
0.073 & 0.037 & 0.217 & 2400 \\
DeepV2D~\cite{deepv2d} &  
0.057 & \textbf{0.010} & \underline{0.168} & 690 \\
NeuralRecon~\cite{Sun_2021_CVPR} & 
\textbf{0.047} & 0.024 & \textbf{0.164} & \textbf{30} \\
\midrule
\textbf{\MethodAcronym} ($128 \times 192$)&
0.059 & 0.022 & 0.184 &  \underline{49} \\
\textbf{\MethodAcronym} ($240 \times 320$) &
\underline{0.056} & \underline{0.019} & 0.176 &  78 \\
\midrule
\midrule
\multicolumn{5}{l}{\textbf{Zero-shot transfer to 7-Scenes}~\cite{shotton2013scene}} \\
\midrule
DeMoN~\cite{ummenhofer2017demon} &  
0.389 & 0.420 & 0.855 & 110 \\
NeuralRGBD~\cite{liu2019neural} &
0.176 & 0.112 & 0.441 & 202 \\
DPSNet~\cite{im2019dpsnet} &  
0.199 & 0.142 & 0.438  & 322 \\
DeepV2D~\cite{deepv2d} &  
0.437 & 0.553 & 0.869  & 347 \\
CNMNet~\cite{long2020occlusion} &   
0.161 & 0.083 & 0.361 & 80 \\
NeuralRecon~\cite{Sun_2021_CVPR} &  
0.155 & 0.104 & 0.347 & \textbf{30} \\
EST~\cite{yifan2021input} &  
\underline{0.118} & \underline{0.052} & \underline{0.298} & 71 \\
\midrule
\textbf{\MethodAcronym} ($128 \times 192$) & 
\textbf{0.100} & \textbf{0.039} & \textbf{0.253} &  \underline{49} \\
\bottomrule
\end{tabular}
\caption{\textbf{Depth estimation results on ScanNet and 7-Scenes}. 
\MethodAcronym is competitive with other state-of-the-art methods on ScanNet, and outperforms all published methods in zero-shot transfer to 7-Scenes by a large margin.
}
\vspace{-8mm}
\label{tab:depth_scannet_temporal}
\end{table}

\vspace{-2mm}
\subsection{Depth from Novel Viewpoints}
\label{sec:depth_synthesis}

We previously discussed the strong performance that \MethodAcronym achieves on traditional depth estimation benchmarks, and showed how it improves out-of-domain generalization by a wide margin. Here, we explore another aspect of generalization that our architecture enables: viewpoint generalization. This is possible because, in addition to traditional depth estimation from RGB images, \MethodAcronym can also generate depth maps from arbitrary viewpoints since it only requires camera embeddings to decode estimates. We explore this capability in two different ways: \emph{interpolation} and \emph{extrapolation}. When interpolating, we encode frames at $\{t-5,t+5\}$, and decode virtual depth maps at locations $\{t-4,\dots,t+4\}$. When extrapolating, we encode frames at $\{t-5,\dots,t-1\}$, and decode virtual depth maps at locations $\{t,\dots,t+8\}$. We use the same training and test splits as in our stereo experiments, with a downsampling factor of $20$ to encourage smaller overlaps between frames. As baselines for comparison, we consider the explicit projection of 3D information from encoded frames onto these new viewpoints. We evaluate both standard depth estimation networks~\cite{monodepth2,packnet,lee2019big} as well as \MethodAcronym itself, that can be used to either explicitly project information from encoded frames onto new viewpoints (projection), or query from the latent representation at that same location (query).

\begin{figure}[t!]
\centering
\subfloat[Depth interpolation results.]{
\label{fig:synthesis_interpolation}
\includegraphics[width=0.48\textwidth,height=3.0cm]{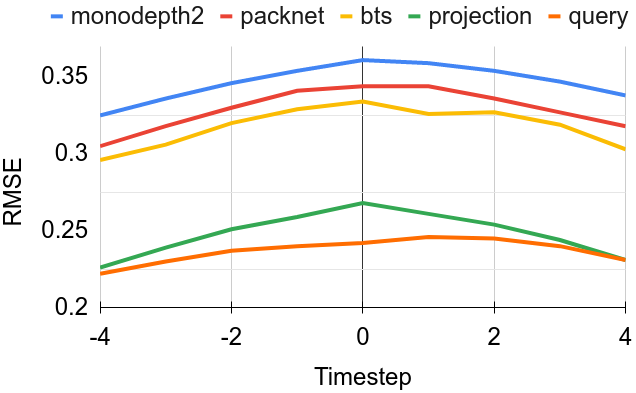}
}
\subfloat[Depth extrapolation results.]{
\label{fig:synthesis_extrapolation}
\includegraphics[width=0.48\textwidth,height=3.0cm]{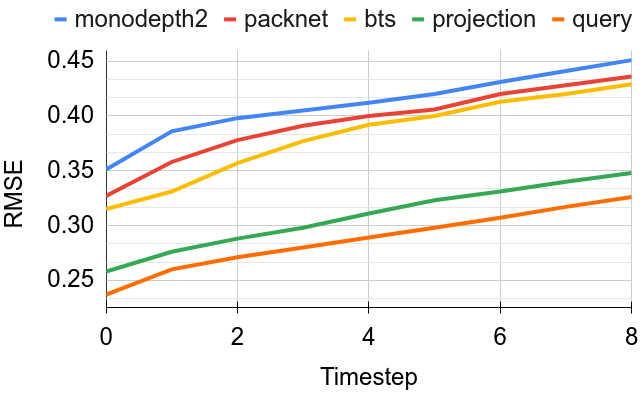}
}
\\ 
\subfloat[Depth extrapolation to future timesteps.  Images and ground-truth depth maps are displayed only for comparison. \MethodAcronym can complete unseen portions of the scene in a geometrically-consistent way, generating dense depth maps from novel viewpoints.]{
\label{fig:synthesis_qualitative}
\includegraphics[width=0.99\textwidth]{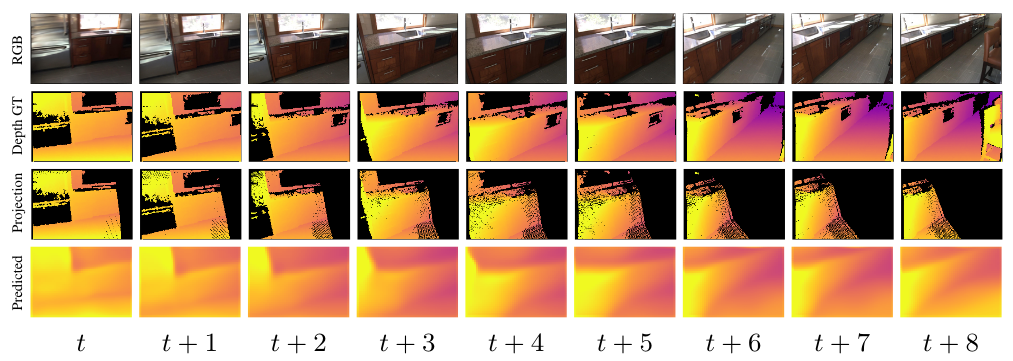}
}
\caption{\textbf{Depth estimation results} from novel viewpoints.}
\label{fig:depth_synthesis}
\vspace{-4mm}
\end{figure}

Figure~\ref{fig:depth_synthesis} reports results in terms of root mean squared error (RMSE) considering only valid projected pixels.  Of particular note, our multi-frame depth estimation architecture significantly outperforms other single-frame baselines. However, and most importantly, results obtained by implicit querying consistently outperform those obtained via explicit projection. This indicates that our model is able to improve upon available encoded information via the learned latent representation. Furthermore, we also generate geometrically consistent estimates for areas without valid explicit projections (Figure~\ref{fig:synthesis_qualitative}).  As the camera tilts to the right, the floor is smoothly recreated in unobserved areas, as well as the partially observed bench. Interestingly, the chair at the extreme right was not recreated, which could be seen as a failure case. However, because the chair was never observed in the first place, it is reasonable for the model to assume the area is empty, and recreate it as a continuation of the floor. 









\section{Conclusion}
We introduced \MethodName (\MethodAcronym), a generalist framework for training multi-view consistent depth estimators. Rather than explicitly enforcing geometric constraints at an architecture or loss level,
we use geometric embeddings to condition network inputs, alongside visual information. To learn a geometrically-consistent latent representation, we propose a series of 3D augmentations designed to promote viewpoint, rotation, and translation equivariance.  
We also show that the introduction of view synthesis as an auxiliary task improves depth estimation without requiring additional ground truth.  
We achieve state-of-the-art results on the popular ScanNet stereo benchmark, and competitive results on the ScanNet video benchmark with no iterative refinement or explicit geometry modeling. We also demonstrate strong generalization properties by achieving state-of-the-art results on zero-shot transfer from ScanNet to 7-Scenes.
The general nature of our framework enables many exciting avenues for future work, including additional tasks such as optical flow, extension to dynamic scenes, spatio-temporal representations, and uncertainty estimation.



\appendix
\section{Implementation Details}

\subsection{Training parameters}

We implemented our models using PyTorch,
with distributed training across eight A100 GPUs. We used grid search to choose training parameters, that include: view synthesis weight $\lambda_s = 5.0$, virtual camera loss weight $\lambda_v=0.5$, virtual camera projection noise $\sigma_v = 0.25$, canonical jittering noise $\sigma_t=\sigma_r=0.1$, and batch size $b=32$ ($4$ per GPU). We use the AdamW optimizer~\cite{loshchilov2019decoupled}, with standard parameters $\beta_1=0.9$, $\beta_2=0.999$, a weight decay of $w=10^{-4}$, and an initial learning rate of $lr=2 \cdot 10^{-4}$. For our stereo experiments, we train for $200$ epochs, halving the learning rate every $80$ epochs. For our video experiments, we train for $100$ epochs, halving the learning rate every $40$ epochs. Higher-resolution fine-tuning is performed for $50$ epochs for stereo experiments, and $10$ epochs for video experiments, with $lr=2 \cdot 10^{-5}$.

\subsection{Architecture Details}

Following recent work~\cite{yifan2021input}, we use $K_o=20$ and $K_r=10$ as the number of Fourier frequencies for camera embeddings, with maximum resolution $\mu_o = \mu_r = 60$. Our encoder embeddings have dimensionality $C_e=960 + 186 = 1146$, due to the use of both image and camera information. Our decoder embeddings have dimensionality $C_d=186$, since only camera information is required to produce estimates. Our latent representation $\mathcal{R}$ is of dimensionality $2048 \times 512$. Input images are resized to $128 \times 192$, and following standard protocol~\cite{deepv2d} output depth maps are compared to ground-truth resized to $480 \times 640$. We use the following hyperparameters for our Perceiver~IO implementation: $1$ block, $1$ input cross-attention, $8$ self-attention layers (with $8$ heads) and $1$ output cross-attention.  Cross attention layers have only $1$ head. We found that larger Perceiver~IO models (i.e., with more blocks, number of heads, and self-/cross-attention layers) did not improve results and significantly increased training time. The latest developments in the Perceiver architecture~\cite{hier-perceiver} could be used to further improve performance and inference speed, and will be considered in future work.

\section{Canonical Jittering Test-time Ablation}

In Section 4.3 of the main text, we ablate the effects of using our proposed data augmentation techniques, designed to improve multi-view consistency in the learned latent representation. In Figure 4b we provide an additional experiment in which we vary the amount of virtual camera noise $\sigma_v$ at train and test time, and show that training at higher noise levels not only improves depth estimation performance at the target location (up to a certain value, of $\sigma_v=0.25$m), but also when decoding estimates from novel viewpoints. 

Here we perform a similar experiment targeting another proposed data augmentation technique: canonical jittering. Two different models were trained, with and without canonical jittering, and both were evaluated under different noise levels at test time.  Note that, while this augmentation does not change scene geometry, it changes the camera embeddings used for encoding and decoding information.
Results are presented in Table \ref{table:pose_jittering}. As we can see, the model trained with canonical jittering not only performs better when evaluating at the target location, but is also more robust to increasing levels of noise at test time. 
\captionsetup[table]{skip=6pt}
\begin{table}[t!]
\renewcommand{\arraystretch}{1.1}
\centering
{
\small
\setlength{\tabcolsep}{0.3em}
\begin{tabular}{|l|cccc|}
\toprule
\diagbox{Train}{Test} & 
$0.0$m  & 
$0.1$m &
$0.2$m &
$0.5$m
\\
\midrule
$0.0$m
&
$0.101$ & $0.202$ & $0.226$ & $0.291$ \\
$0.1$m
&
$0.097$ & $0.160$ & $0.184$ & $0.242$ \\ 
\bottomrule
\end{tabular}
}
\caption{\textbf{Effects of canonical jittering at train and test time}. The model trained with canonical jittering ($\sigma_t = \sigma_r = 0.1$m) not only performs better when evaluated at the target location ($\sigma_t = \sigma_r = 0.0$m), but is also more robust to different levels of canonical jittering at test time. The results shown are average Abs. Rel. of the two predicted stereo depths maps, without ground-truth scaling.}
\vspace{-6mm}
\label{table:pose_jittering}
\end{table}


\section{Higher Resolution Fine-Tuning}
One of the main challenges of training Transformer-based architectures has been the $O(N^2)$ self-attention memory scaling with input size. 
This means that the resolution of recent models has been fairly limited (e.g. the view synthesis model of Sajjadi et al.~\cite{sajjadi2021scene} primarily trains on low-resolution images, with a highest resolution of $128 \times 176$), hindering their application to real-world scenes.  
Perceiver~IO decouples input resolution from the the learned latent representation, which enables training and real-time inference at higher resolutions~\cite{yifan2021input}.
In our experiments, we found it advantageous to train using a resolution curriculum, first at a lower resolution ($128 \times 192$), and then fine-tune at a higher resolution $(240 \times 320)$. Note that, because the camera parameters are also scaled to the proper resolution, the scene geometry does not change, only (a) the number of embeddings generated per camera, and (b) the image embeddings, since resolution changes image features.  Thus, training at lower resolutions enables the faster learning of our desired multi-view latent scene representation, which can then be fine-tuned at higher resolutions for further improvements.
As an alternative, we also experimented with the strategy of \textit{sampling} rays at higher resolution (similar to NeRF~\cite{mildenhall2020nerf} and SRT~\cite{sajjadi2021scene}). However, we found that this approach led to unstable training and longer convergence times. As future work, we plan to investigate how training and and inference can be scaled up to even higher resolutions. 

\begin{figure}[t!]
\captionsetup[subfloat]{labelformat=empty}
\subfloat{\includegraphics[width=0.16\textwidth,height=1.4cm]{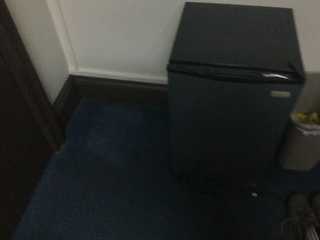}} \!
\subfloat{\includegraphics[width=0.16\textwidth,height=1.4cm]{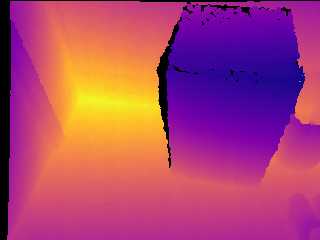}} \!
\subfloat{\includegraphics[width=0.17\textwidth,height=1.4cm]{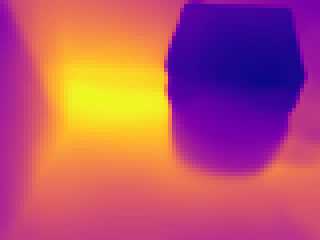}}  \!
\subfloat{\includegraphics[width=0.17\textwidth,height=1.4cm]{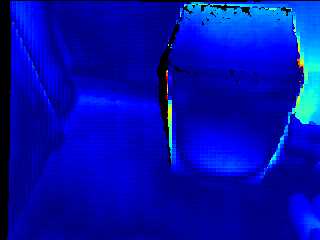}} \! 
\subfloat{\includegraphics[width=0.16\textwidth,height=1.4cm]{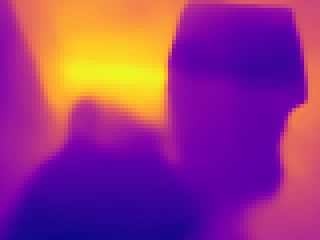}}  \!
\subfloat{\includegraphics[width=0.16\textwidth,height=1.4cm]{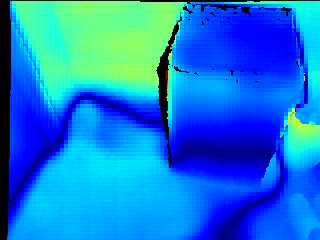}}
\\
\subfloat{\includegraphics[width=0.16\textwidth,height=1.4cm]{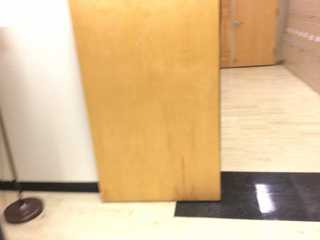}} \!
\subfloat{\includegraphics[width=0.16\textwidth,height=1.4cm]{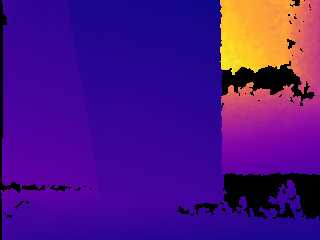}} \!
\subfloat{\includegraphics[width=0.17\textwidth,height=1.4cm]{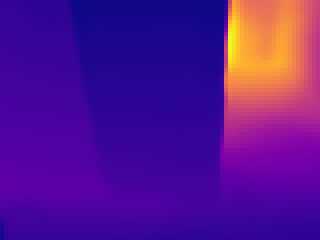}}  \!
\subfloat{\includegraphics[width=0.17\textwidth,height=1.4cm]{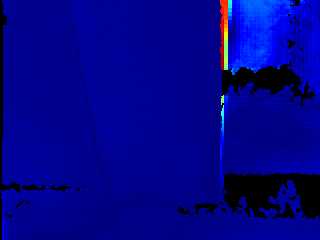}} \!
\subfloat{\includegraphics[width=0.16\textwidth,height=1.4cm]{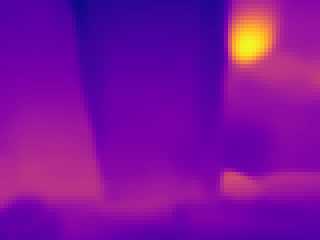}}  \!
\subfloat{\includegraphics[width=0.16\textwidth,height=1.4cm]{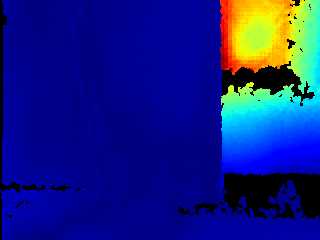}}
\\ 
\subfloat{\includegraphics[width=0.16\textwidth,height=1.4cm]{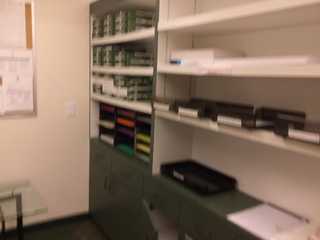}} \!
\subfloat{\includegraphics[width=0.16\textwidth,height=1.4cm]{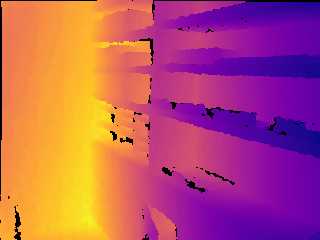}} \!
\subfloat{\includegraphics[width=0.17\textwidth,height=1.4cm]{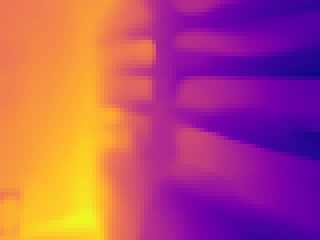}}  \!
\subfloat{\includegraphics[width=0.17\textwidth,height=1.4cm]{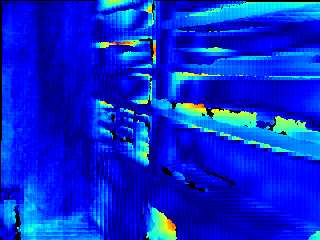}}  \!
\subfloat{\includegraphics[width=0.16\textwidth,height=1.4cm]{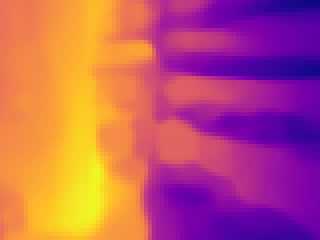}}  \!
\subfloat{\includegraphics[width=0.16\textwidth,height=1.4cm]{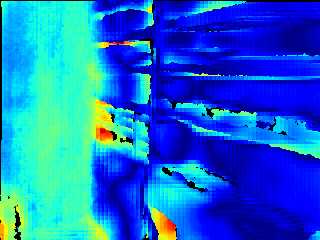}}
\\ 
\subfloat[Input Image]{\includegraphics[width=0.16\textwidth,height=1.4cm]{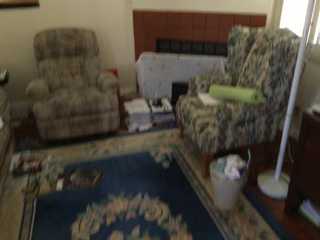}} \!
\subfloat[GT depth]{\includegraphics[width=0.16\textwidth,height=1.4cm]{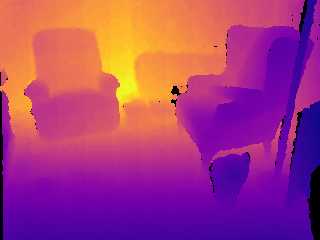}} \!
\subfloat[DeFiNe depth]{\includegraphics[width=0.17\textwidth,height=1.4cm]{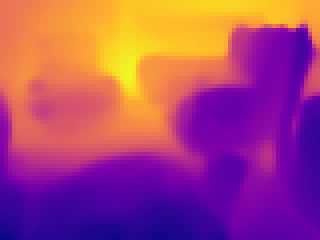}} \!
\subfloat[DeFiNe error]{\includegraphics[width=0.17\textwidth,height=1.4cm]{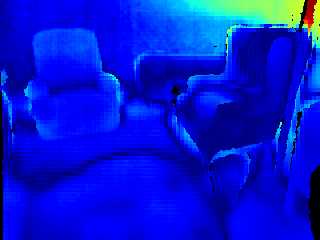}} \!
\subfloat[IIB depth]{\includegraphics[width=0.16\textwidth,height=1.4cm]{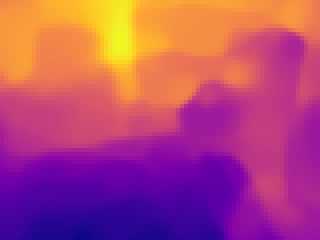}} \!
\subfloat[IIB error]{\includegraphics[width=0.16\textwidth,height=1.4cm]{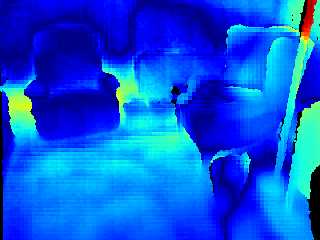}}
\caption{\textbf{Qualitative comparison of DeFiNe} relative to the IIB~\cite{yifan2021input} baseline.  
Our architecture improves depth estimation quality in (i) smooth and textureless areas, (ii) far away regions, and (iii) image boundaries and depth discontinuities.
}
\label{fig:iib}
\vspace{-5mm}
\end{figure}

\section{Comparison to IIB}
IIB~\cite{yifan2021input} is a recently proposed stereo depth estimation method that also uses a Perceiver~IO-based architecture.  Their major contribution is a geometrically-motivated epipolar inductive bias to encourage multi-view consistency. In Table 1 and Figure 4 of the main text, we show that our DeFiNe architecture significantly improves over the IIB baseline on the ScanNet-Stereo benchmark ($0.116$ vs.\ $0.089$ Abs.\ Rel.).  
Given that code and pre-trained models to replicate the IIB results are not available, we trained a model following the instructions in~\cite{yifan2021input}, achieving similar performance as reported in the paper. 

Some qualitative examples from this model are depicted in Figure \ref{fig:iib}, as well as examples from our DeFiNe architecture. As we can see, our proposed 3D augmentations and joint view synthesis learning also lead to significant qualitative improvements over IIB results. In particular, we consistently perform better in smooth and textureless areas, as well as far away regions and depth discontinuities. We attribute this behavior to an increase in scene diversity due to our contributions, that enables the learning of a more consistent multi-view latent scene representation.



\section{Depth from Novel Viewpoints}

\begin{figure}[t!]
\captionsetup[subfloat]{labelformat=empty}
\vspace{2mm} 
\rotatebox{90}{\hspace{1.0mm} \tiny{RGB}} \!
\subfloat{\includegraphics[width=0.1\textwidth,height=0.85cm]{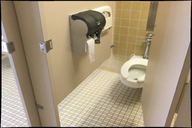}} \hspace{0.1mm}
\subfloat{\includegraphics[width=0.1\textwidth,height=0.85cm]{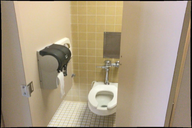}} \hspace{0.1mm}
\subfloat{\includegraphics[width=0.1\textwidth,height=0.85cm]{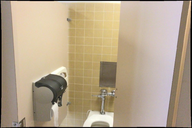}} \hspace{0.1mm}
\subfloat{\includegraphics[width=0.1\textwidth,height=0.85cm]{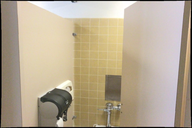}} \hspace{0.1mm}
\subfloat{\includegraphics[width=0.1\textwidth,height=0.85cm]{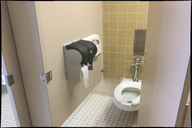}} \hspace{0.1mm}
\subfloat{\includegraphics[width=0.1\textwidth,height=0.85cm]{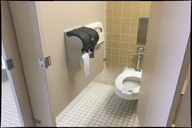}} \hspace{0.1mm}
\subfloat{\includegraphics[width=0.1\textwidth,height=0.85cm]{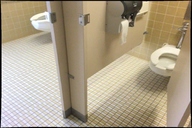}} \hspace{0.1mm}
\subfloat{\includegraphics[width=0.1\textwidth,height=0.85cm]{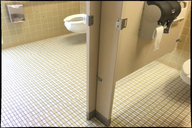}} \hspace{0.1mm}
\subfloat{\includegraphics[width=0.1\textwidth,height=0.85cm]{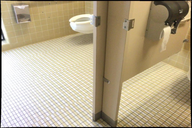}} 
\\ 
\rotatebox{90}{\hspace{1.0mm} \tiny{GT}} \!
\subfloat{\includegraphics[width=0.1\textwidth,height=0.85cm]{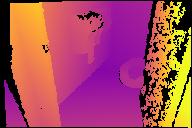}} \hspace{0.1mm}
\subfloat{\includegraphics[width=0.1\textwidth,height=0.85cm]{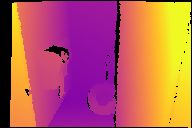}} \hspace{0.1mm}
\subfloat{\includegraphics[width=0.1\textwidth,height=0.85cm]{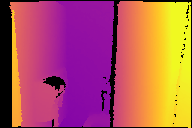}} \hspace{0.1mm}
\subfloat{\includegraphics[width=0.1\textwidth,height=0.85cm]{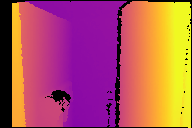}} \hspace{0.1mm}
\subfloat{\includegraphics[width=0.1\textwidth,height=0.85cm]{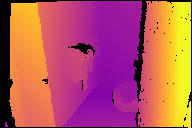}} \hspace{0.1mm}
\subfloat{\includegraphics[width=0.1\textwidth,height=0.85cm]{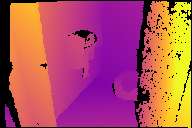}} \hspace{0.1mm}
\subfloat{\includegraphics[width=0.1\textwidth,height=0.85cm]{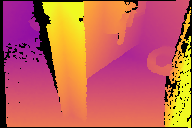}} \hspace{0.1mm}
\subfloat{\includegraphics[width=0.1\textwidth,height=0.85cm]{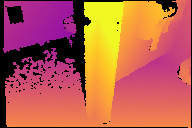}} \hspace{0.1mm}
\subfloat{\includegraphics[width=0.1\textwidth,height=0.85cm]{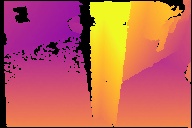}} 
\\ 
\rotatebox{90}{\hspace{0.4mm} \tiny{Proj.}}
\subfloat{\includegraphics[width=0.1\textwidth,height=0.85cm]{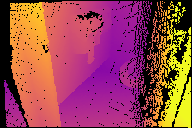}} \hspace{0.1mm}
\subfloat{\includegraphics[width=0.1\textwidth,height=0.85cm]{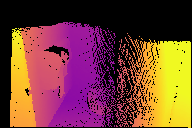}} \hspace{0.1mm}
\subfloat{\includegraphics[width=0.1\textwidth,height=0.85cm]{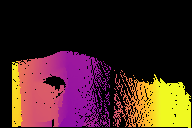}} \hspace{0.1mm}
\subfloat{\includegraphics[width=0.1\textwidth,height=0.85cm]{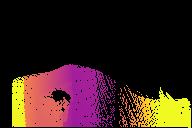}} \hspace{0.1mm}
\subfloat{\includegraphics[width=0.1\textwidth,height=0.85cm]{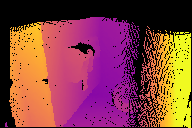}} \hspace{0.1mm}
\subfloat{\includegraphics[width=0.1\textwidth,height=0.85cm]{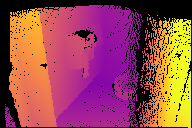}} \hspace{0.1mm}
\subfloat{\includegraphics[width=0.1\textwidth,height=0.85cm]{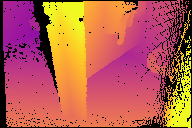}} \hspace{0.1mm}
\subfloat{\includegraphics[width=0.1\textwidth,height=0.85cm]{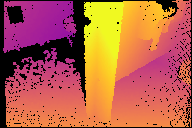}} \hspace{0.1mm}
\subfloat{\includegraphics[width=0.1\textwidth,height=0.85cm]{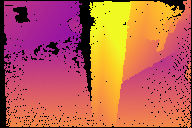}} 
\\ 
\rotatebox{90}{\hspace{0.1mm} \tiny{Pred.}} \!
\subfloat[$t$]{\includegraphics[width=0.1\textwidth,height=0.85cm]{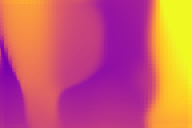}} \hspace{0.1mm}
\subfloat[$t+1$]{\includegraphics[width=0.1\textwidth,height=0.85cm]{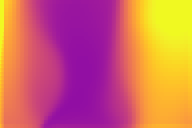}} \hspace{0.1mm}
\subfloat[$t+2$]{\includegraphics[width=0.1\textwidth,height=0.85cm]{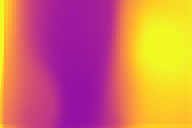}} \hspace{0.1mm}
\subfloat[$t+3$]{\includegraphics[width=0.1\textwidth,height=0.85cm]{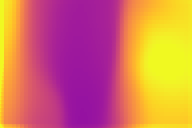}} \hspace{0.1mm}
\subfloat[$t+4$]{\includegraphics[width=0.1\textwidth,height=0.85cm]{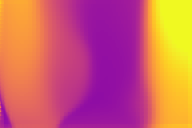}} \hspace{0.1mm}
\subfloat[$t+5$]{\includegraphics[width=0.1\textwidth,height=0.85cm]{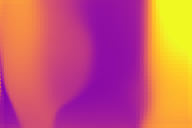}} \hspace{0.1mm}
\subfloat[$t+6$]{\includegraphics[width=0.1\textwidth,height=0.85cm]{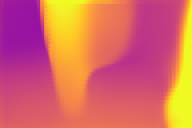}} \hspace{0.1mm}
\subfloat[$t+7$]{\includegraphics[width=0.1\textwidth,height=0.85cm]{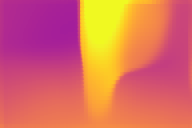}} \hspace{0.1mm}
\subfloat[$t+8$]{\includegraphics[width=0.1\textwidth,height=0.85cm]{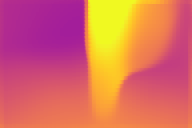}}
\vspace{2mm} \\ 
\rotatebox{90}{\hspace{1.0mm} \tiny{RGB}} \!
\subfloat{\includegraphics[width=0.1\textwidth,height=0.85cm]{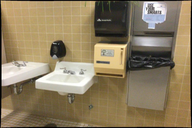}} \hspace{0.1mm}
\subfloat{\includegraphics[width=0.1\textwidth,height=0.85cm]{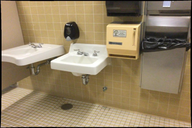}} \hspace{0.1mm}
\subfloat{\includegraphics[width=0.1\textwidth,height=0.85cm]{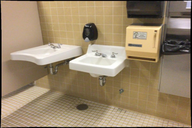}} \hspace{0.1mm}
\subfloat{\includegraphics[width=0.1\textwidth,height=0.85cm]{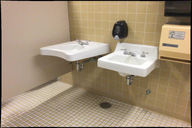}} \hspace{0.1mm}
\subfloat{\includegraphics[width=0.1\textwidth,height=0.85cm]{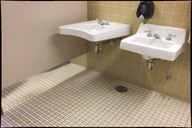}} \hspace{0.1mm}
\subfloat{\includegraphics[width=0.1\textwidth,height=0.85cm]{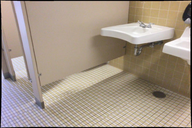}} \hspace{0.1mm}
\subfloat{\includegraphics[width=0.1\textwidth,height=0.85cm]{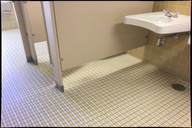}} \hspace{0.1mm}
\subfloat{\includegraphics[width=0.1\textwidth,height=0.85cm]{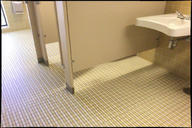}} \hspace{0.1mm}
\subfloat{\includegraphics[width=0.1\textwidth,height=0.85cm]{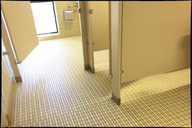}} 
\\ 
\rotatebox{90}{\hspace{1.0mm} \tiny{GT}} \!
\subfloat{\includegraphics[width=0.1\textwidth,height=0.85cm]{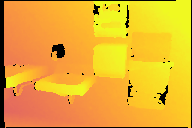}} \hspace{0.1mm}
\subfloat{\includegraphics[width=0.1\textwidth,height=0.85cm]{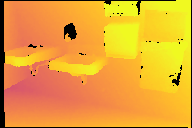}} \hspace{0.1mm}
\subfloat{\includegraphics[width=0.1\textwidth,height=0.85cm]{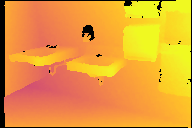}} \hspace{0.1mm}
\subfloat{\includegraphics[width=0.1\textwidth,height=0.85cm]{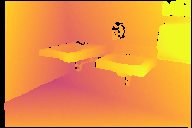}} \hspace{0.1mm}
\subfloat{\includegraphics[width=0.1\textwidth,height=0.85cm]{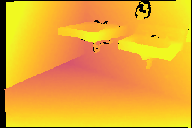}} \hspace{0.1mm}
\subfloat{\includegraphics[width=0.1\textwidth,height=0.85cm]{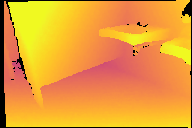}} \hspace{0.1mm}
\subfloat{\includegraphics[width=0.1\textwidth,height=0.85cm]{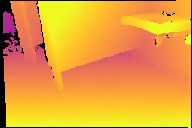}} \hspace{0.1mm}
\subfloat{\includegraphics[width=0.1\textwidth,height=0.85cm]{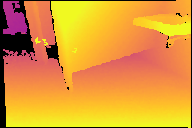}} \hspace{0.1mm}
\subfloat{\includegraphics[width=0.1\textwidth,height=0.85cm]{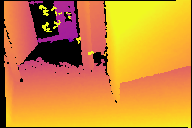}} 
\\ 
\rotatebox{90}{\hspace{0.4mm} \tiny{Proj.}}
\subfloat{\includegraphics[width=0.1\textwidth,height=0.85cm]{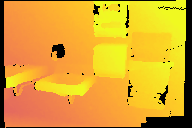}} \hspace{0.1mm}
\subfloat{\includegraphics[width=0.1\textwidth,height=0.85cm]{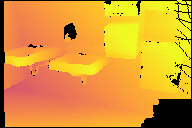}} \hspace{0.1mm}
\subfloat{\includegraphics[width=0.1\textwidth,height=0.85cm]{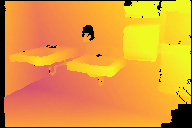}} \hspace{0.1mm}
\subfloat{\includegraphics[width=0.1\textwidth,height=0.85cm]{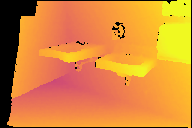}} \hspace{0.1mm}
\subfloat{\includegraphics[width=0.1\textwidth,height=0.85cm]{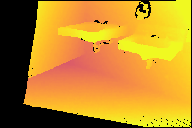}} \hspace{0.1mm}
\subfloat{\includegraphics[width=0.1\textwidth,height=0.85cm]{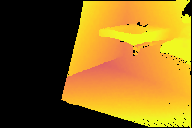}} \hspace{0.1mm}
\subfloat{\includegraphics[width=0.1\textwidth,height=0.85cm]{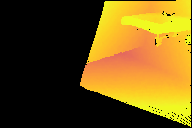}} \hspace{0.1mm}
\subfloat{\includegraphics[width=0.1\textwidth,height=0.85cm]{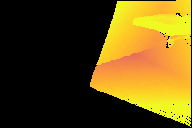}} \hspace{0.1mm}
\subfloat{\includegraphics[width=0.1\textwidth,height=0.85cm]{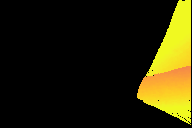}} 
\\ 
\rotatebox{90}{\hspace{0.1mm} \tiny{Pred.}} \!
\subfloat[$t$]{\includegraphics[width=0.1\textwidth,height=0.85cm]{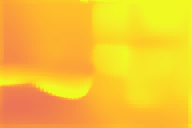}} \hspace{0.1mm}
\subfloat[$t+1$]{\includegraphics[width=0.1\textwidth,height=0.85cm]{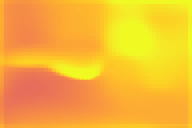}} \hspace{0.1mm}
\subfloat[$t+2$]{\includegraphics[width=0.1\textwidth,height=0.85cm]{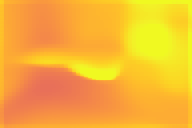}} \hspace{0.1mm}
\subfloat[$t+3$]{\includegraphics[width=0.1\textwidth,height=0.85cm]{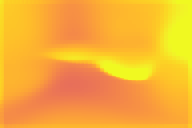}} \hspace{0.1mm}
\subfloat[$t+4$]{\includegraphics[width=0.1\textwidth,height=0.85cm]{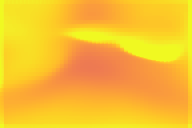}} \hspace{0.1mm}
\subfloat[$t+5$]{\includegraphics[width=0.1\textwidth,height=0.85cm]{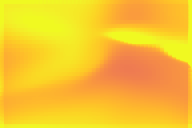}} \hspace{0.1mm}
\subfloat[$t+6$]{\includegraphics[width=0.1\textwidth,height=0.85cm]{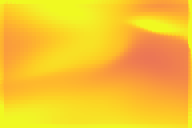}} \hspace{0.1mm}
\subfloat[$t+7$]{\includegraphics[width=0.1\textwidth,height=0.85cm]{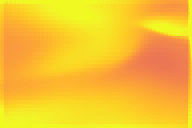}} \hspace{0.1mm}
\subfloat[$t+8$]{\includegraphics[width=0.1\textwidth,height=0.85cm]{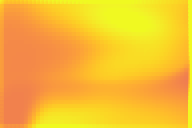}}
\vspace{2mm} \\ 
\rotatebox{90}{\hspace{1.0mm} \tiny{RGB}} \!
\subfloat{\includegraphics[width=0.1\textwidth,height=0.85cm]{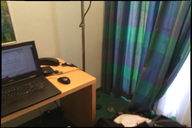}} \hspace{0.1mm}
\subfloat{\includegraphics[width=0.1\textwidth,height=0.85cm]{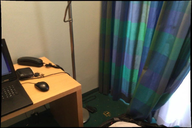}} \hspace{0.1mm}
\subfloat{\includegraphics[width=0.1\textwidth,height=0.85cm]{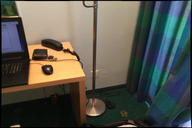}} \hspace{0.1mm}
\subfloat{\includegraphics[width=0.1\textwidth,height=0.85cm]{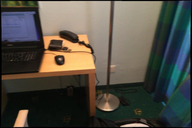}} \hspace{0.1mm}
\subfloat{\includegraphics[width=0.1\textwidth,height=0.85cm]{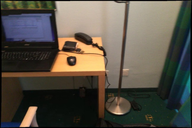}} \hspace{0.1mm}
\subfloat{\includegraphics[width=0.1\textwidth,height=0.85cm]{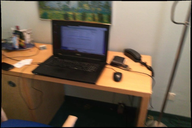}} \hspace{0.1mm}
\subfloat{\includegraphics[width=0.1\textwidth,height=0.85cm]{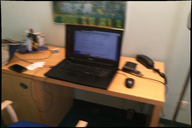}} \hspace{0.1mm}
\subfloat{\includegraphics[width=0.1\textwidth,height=0.85cm]{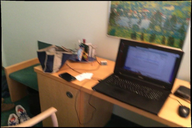}} \hspace{0.1mm}
\subfloat{\includegraphics[width=0.1\textwidth,height=0.85cm]{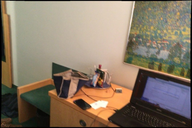}} 
\\ 
\rotatebox{90}{\hspace{1.0mm} \tiny{GT}} \!
\subfloat{\includegraphics[width=0.1\textwidth,height=0.85cm]{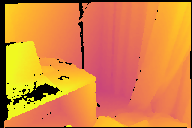}} \hspace{0.1mm}
\subfloat{\includegraphics[width=0.1\textwidth,height=0.85cm]{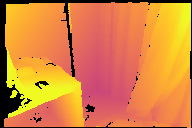}} \hspace{0.1mm}
\subfloat{\includegraphics[width=0.1\textwidth,height=0.85cm]{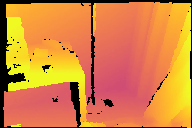}} \hspace{0.1mm}
\subfloat{\includegraphics[width=0.1\textwidth,height=0.85cm]{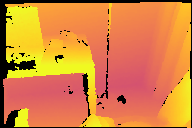}} \hspace{0.1mm}
\subfloat{\includegraphics[width=0.1\textwidth,height=0.85cm]{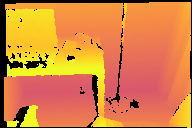}} \hspace{0.1mm}
\subfloat{\includegraphics[width=0.1\textwidth,height=0.85cm]{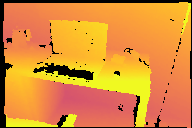}} \hspace{0.1mm}
\subfloat{\includegraphics[width=0.1\textwidth,height=0.85cm]{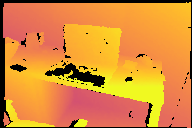}} \hspace{0.1mm}
\subfloat{\includegraphics[width=0.1\textwidth,height=0.85cm]{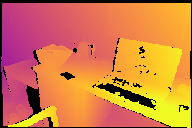}} \hspace{0.1mm}
\subfloat{\includegraphics[width=0.1\textwidth,height=0.85cm]{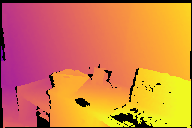}} 
\\ 
\rotatebox{90}{\hspace{0.4mm} \tiny{Proj.}}
\subfloat{\includegraphics[width=0.1\textwidth,height=0.85cm]{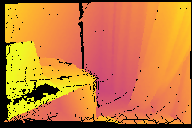}} \hspace{0.1mm}
\subfloat{\includegraphics[width=0.1\textwidth,height=0.85cm]{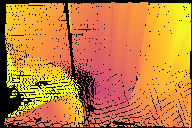}} \hspace{0.1mm}
\subfloat{\includegraphics[width=0.1\textwidth,height=0.85cm]{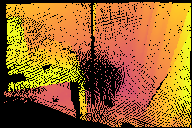}} \hspace{0.1mm}
\subfloat{\includegraphics[width=0.1\textwidth,height=0.85cm]{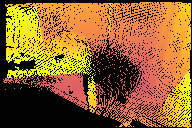}} \hspace{0.1mm}
\subfloat{\includegraphics[width=0.1\textwidth,height=0.85cm]{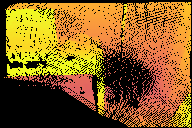}} \hspace{0.1mm}
\subfloat{\includegraphics[width=0.1\textwidth,height=0.85cm]{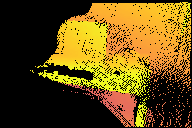}} \hspace{0.1mm}
\subfloat{\includegraphics[width=0.1\textwidth,height=0.85cm]{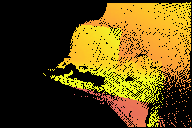}} \hspace{0.1mm}
\subfloat{\includegraphics[width=0.1\textwidth,height=0.85cm]{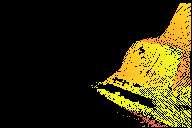}} \hspace{0.1mm}
\subfloat{\includegraphics[width=0.1\textwidth,height=0.85cm]{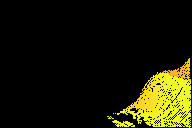}} 
\\ 
\rotatebox{90}{\hspace{0.1mm} \tiny{Pred.}} \!
\subfloat[$t$]{\includegraphics[width=0.1\textwidth,height=0.85cm]{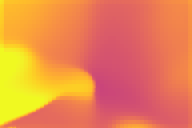}} \hspace{0.1mm}
\subfloat[$t+1$]{\includegraphics[width=0.1\textwidth,height=0.85cm]{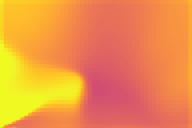}} \hspace{0.1mm}
\subfloat[$t+2$]{\includegraphics[width=0.1\textwidth,height=0.85cm]{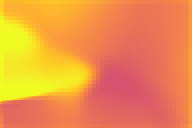}} \hspace{0.1mm}
\subfloat[$t+3$]{\includegraphics[width=0.1\textwidth,height=0.85cm]{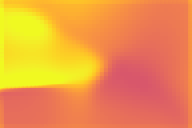}} \hspace{0.1mm}
\subfloat[$t+4$]{\includegraphics[width=0.1\textwidth,height=0.85cm]{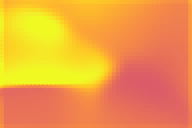}} \hspace{0.1mm}
\subfloat[$t+5$]{\includegraphics[width=0.1\textwidth,height=0.85cm]{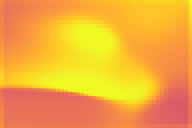}} \hspace{0.1mm}
\subfloat[$t+6$]{\includegraphics[width=0.1\textwidth,height=0.85cm]{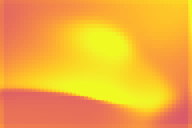}} \hspace{0.1mm}
\subfloat[$t+7$]{\includegraphics[width=0.1\textwidth,height=0.85cm]{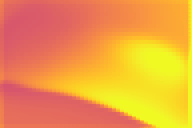}} \hspace{0.1mm}
\subfloat[$t+8$]{\includegraphics[width=0.1\textwidth,height=0.85cm]{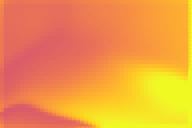}}
\caption{\textbf{ScanNet depth extrapolation examples}, using DeFiNe. In each example, image and camera information from frames at $[t-5,\dots,t-1]$ is encoded, and depth maps corresponding to camera locations at $[t,\dots,t+8]$ are decoded, using only camera information. For each timestep, we show sparse projected ground-truth depth maps (third row), and dense predicted depth maps (fourth row). Our DeFiNe architecture is able to extrapolate from encoded information to fill in missing parts of the scene.}
\label{fig:extra_additional}
\end{figure}

In Table \ref{tab:intra_extra} we provide numerical values to complement our depth interpolation and extrapolation experiments (Figures 7a and 7b from the main text). These experiments show that querying from our learned latent representation improves over the explicit projection of information from encoded views, while also enabling the estimation of dense depth maps from novel viewpoints. Similarly, in Figure \ref{fig:extra_additional} we provide additional qualitative examples of depth extrapolation to future timesteps, showing how DeFiNe can reconstruct unseen portions of the environment in a geometrically-consistent way.

\begin{table*}[t!]
\renewcommand{\arraystretch}{1.0}
\centering
{
\small
\setlength{\tabcolsep}{0.3em}
\subfloat[Depth interpolation results.  
Frames at $\{t-5,t+5\}$ are encoded, and depth maps corresponding to camera locations at $\{t-4,\dots,t+4\}$ are decoded.
]{
\begin{tabular}{l|ccccccccc}
\toprule
Timestep & $-4$ & $-3$ & $-2$ & $-1$ & $0$ & $+1$ & $+2$ & $+3$ & $+4$ \\
\midrule
\% valid pixels & $77.7$ & $68.5$ & $62.6$ & $59.5$ & $58.2$ & $58.7$ & $61.2$ & $66.6$ & $75.9$ \\
\toprule
Monodepth2~\cite{monodepth2} & 0.325 & 0.336 & 0.346 & 0.354 & 0.361 & 0.359 & 0.354 & 0.347 & 0.338 \\
PackNet~\cite{packnet}    & 0.305 & 0.318 & 0.330 & 0.341 & 0.344 & 0.344 & 0.336 & 0.327 & 0.338 \\
BTS~\cite{lee2019big}        & 0.296 & 0.306 & 0.320 & 0.329 & 0.334 & 0.326 & 0.327 & 0.319 & 0.303 \\
\midrule
DeFiNe (projection) & 0.226 & 0.239 & 0.251 & 0.259 & 0.268 & 0.261 & 0.254 & 0.244 & 0.231 \\
DeFiNe (query)      & 0.222 & 0.230 & 0.237 & 0.240 & 0.242 & 0.246 & 0.245 & 0.240 & 0.231 \\
\midrule
\midrule
DeFiNe (query, all) & 0.361 & 0.381 & 0.398 & 0.408 & 0.412 & 0.413 & 0.406 & 0.390 & 0.369 \\
\bottomrule
\end{tabular}
} \\
\subfloat[Depth extrapolation results. 
Frames at $\{t-5,\dots,t-1\}$ are encoded, and depth maps corresponding to camera locations at $\{t,\dots,t+8\}$ are decoded.
]{
\begin{tabular}{l|ccccccccc}
\toprule
Timestep & $0$ & $1$ & $2$ & $3$ & $4$ & $5$ & $6$ & $7$ & $8$ \\
\midrule
\% valid pixels & $91.0$ & $76.1$ & $64.5$ & $55.9$ & $49.6$ & $45.2$ & $42.0$ & $39.5$ & $36.0$ \\
\toprule
Monodepth2~\cite{monodepth2} & 0.351 & 0.386 & 0.398 & 0.405 & 0.412 & 0.420 & 0.431 & 0.441 & 0.453 \\
PackNet~\cite{packnet} & 0.327 & 0.358 & 0.378 & 0.391 & 0.400 & 0.406 & 0.420 & 0.428 & 0.436 \\
BTS~\cite{lee2019big} & 0.315 & 0.331 & 0.357 & 0.377 & 0.392 & 0.401 & 0.413 & 0.424 & 0.429 \\
\midrule
DeFiNe (projection) & 0.258 & 0.276 & 0.288 & 0.298 & 0.311 & 0.323 & 0.331 & 0.340 & 0.348 \\
DeFiNe (query) & 0.237 & 0.260 & 0.271 & 0.280 & 0.289 & 0.298 & 0.307 & 0.317 & 0.326 \\
\midrule
\midrule
DeFiNe (query, all) & 0.326 & 0.370 & 0.405 & 0.438 & 0.468 & 0.495 & 0.520 & 0.543 & 0.563 \\
\bottomrule
\end{tabular}
}
}
\caption{\textbf{Depth interpolation and extrapolation results}, on ScanNet (complementary to Figures 7a and 7b of the main text). On valid projected pixels, DeFiNe (query) outperforms the explicit projection of all considered single-frame baselines, and it also outperforms the explicit projection of its own estimates, obtained from encoded views (projection). Furthermore, it also enables the estimation of dense depth maps from novel viewpoints, which can be compared to the corresponding ground-truth from that location (query, all).} 
\label{tab:intra_extra}
\vspace{-5mm}
\end{table*}



\clearpage
\bibliographystyle{splncs04}
\bibliography{references}
\end{document}